\documentclass[journal, twocolumn, 10pt]{IEEEtran}

\usepackage{url}            %
\usepackage{booktabs}       %
\usepackage{amsfonts}       %
\usepackage{nicefrac}       %
\usepackage{microtype}      %
\usepackage{hhline}
\usepackage{makecell}
\usepackage{pifont}
\usepackage{mathtools}
\usepackage{natbib}
\setcitestyle{numbers,square}

\usepackage{graphicx} %
\usepackage{subfigure}
\usepackage{amsmath}
\usepackage{amsthm}
\usepackage{amssymb}
\usepackage{bbm}
\usepackage{tikz}
\usepackage{xcolor}
\usetikzlibrary{arrows}

\usepackage{enumitem}
\setlist[enumerate]{itemsep=0mm}
\allowdisplaybreaks

\usepackage{mathrsfs}

\usepackage{cite}
\usepackage{times}
\usepackage{graphicx}
\usepackage{amssymb, amsmath, amsthm}          
{
	\theoremstyle{plain}

}
\usepackage{graphicx}
\usepackage{inputenc}
\usepackage{multirow}
\usepackage{makecell}
\usepackage{enumerate}
\usepackage{verbatim}
\usepackage{hyperref}
\usepackage{url}
\usepackage{booktabs}
\usepackage{algorithm,algpseudocode}
\usepackage{geometry}
\usepackage{fullpage}
\usepackage{lscape}
\usepackage{fancyvrb}
\usepackage{parskip}
\usepackage{enumerate}
\usepackage{framed}
\usepackage{fancyhdr}
\usepackage{tikz}
\usepackage{relsize}
\usepackage{anyfontsize}
\usepackage{pgfplots}
\pgfplotsset{compat=1.17}
\usepackage{diagbox}
\usepackage{mathptmx}
\usepackage{caption}

\input{math_commands}

\begin{document}

\title{Reinforcement Learning With Sparse-executing Action via Sparsity Regularization}

\author{\textbf{Jing-Cheng Pang, Tian Xu, Shengyi Jiang, Yu-Ren Liu and Yang Yu$^\diamond$\thanks{$\diamond$ Corresponding author.}} \\ 
National Key Laboratory for Novel Software Technology, Nanjing University, China \& \\ School of Artificial Intelligence, Nanjing University, China
}

\definecolor{skyblue}{RGB}{65, 116, 214}

\twocolumn
\maketitle

\begin{abstract}
Reinforcement learning (RL) has demonstrated impressive performance in decision-making tasks like embodied control, autonomous driving and financial trading. 
In many decision-making tasks, the agents often encounter the problem of executing actions under limited budgets.
However, classic RL methods typically overlook the challenges posed by such sparse-executing actions. They operate under the assumption that all actions can be taken for a unlimited number of times, both in the formulation of the problem and in the development of effective algorithms.
To tackle the issue of limited action execution in RL, this paper first formalizes the problem as a Sparse Action Markov Decision Process (SA-MDP), in which specific actions in the action space can only be executed for a limited time. Then, we propose a policy optimization algorithm, Action Sparsity REgularization (\methodname), which adaptively handles each action with a distinct preference. \methodname~operates through two steps: First, \methodname~evaluates action sparsity by constrained action sampling. Following this, \methodname~incorporates the sparsity evaluation into policy learning by way of an action distribution regularization. We provide theoretical identification that validates the convergence of \methodname~to a regularized optimal value function.
Experiments on tasks with known sparse-executing actions, where classical RL algorithms struggle to train policy efficiently, \methodname~effectively constrains the action sampling and outperforms baselines. Moreover, we present that \methodname~can generally improve the performance in Atari games, demonstrating its broad applicability.
\end{abstract}


\section{Introduction}
\label{sec:intro}
Reinforcement learning (RL) has proven effective in solving a variety of decision-making problems, including embodied control \citep{embodied_RL,embodied_RL2,RL_robot_control_1,robot_o3f,robot_hidil}, autonomous driving \citep{auto_driving,auto_driving2,auto_driving3} and game playing \citep{RL_play_game_1,RL_play_game_2,RL_play_game_3,RL_play_game_4}. In RL, an agent interacts with the environment and collects samples to optimize a policy. Abundant RL algorithms have been proposed for policy optimization \citep{PPO,policy_gradient,remert_optimization,mg_optimization,talar}. 

\begin{figure}[t]

\centering
\subfigure{
\includegraphics[width=0.45\linewidth]{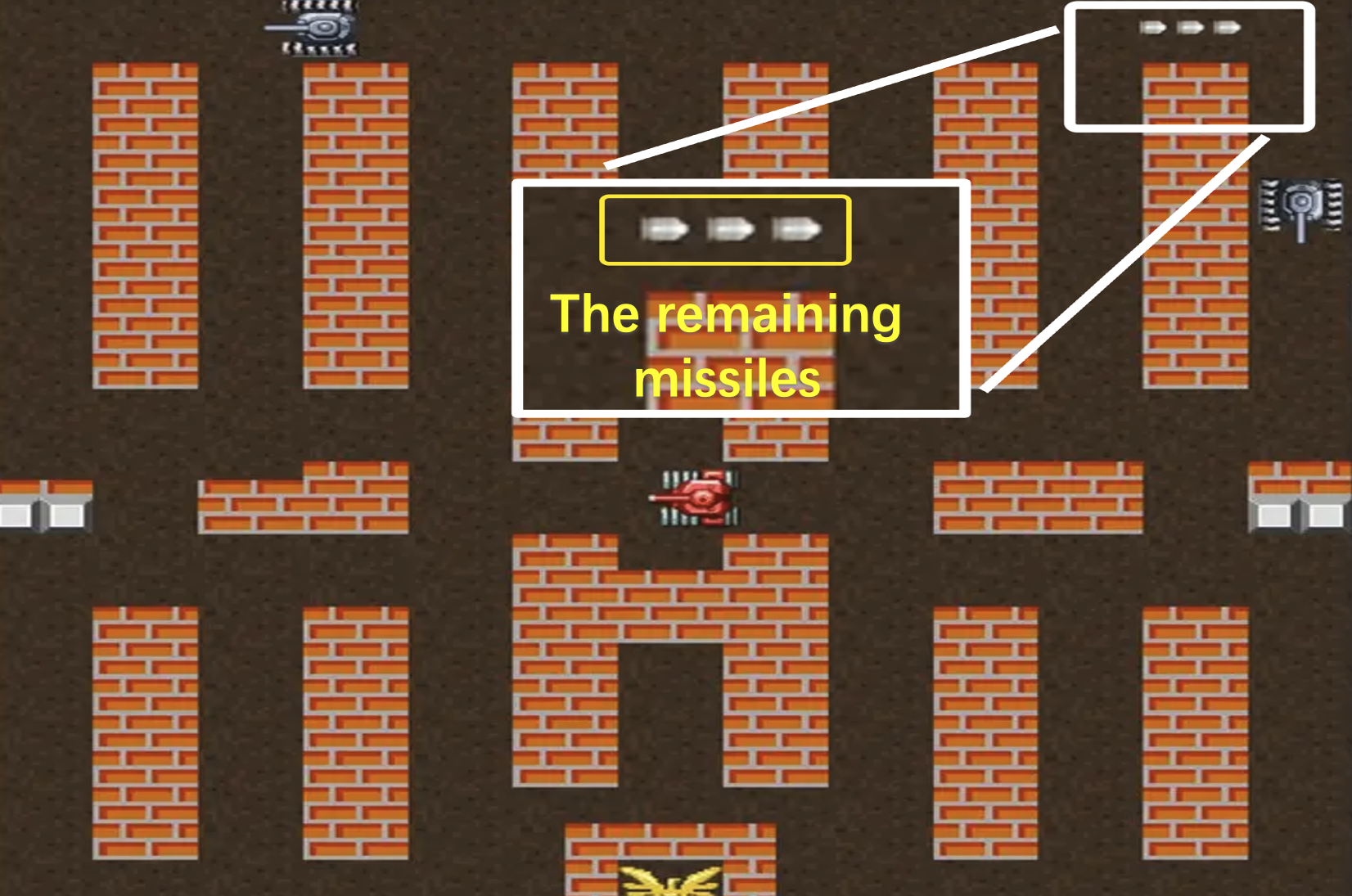}
}
\subfigure{
\includegraphics[width=0.45\linewidth]{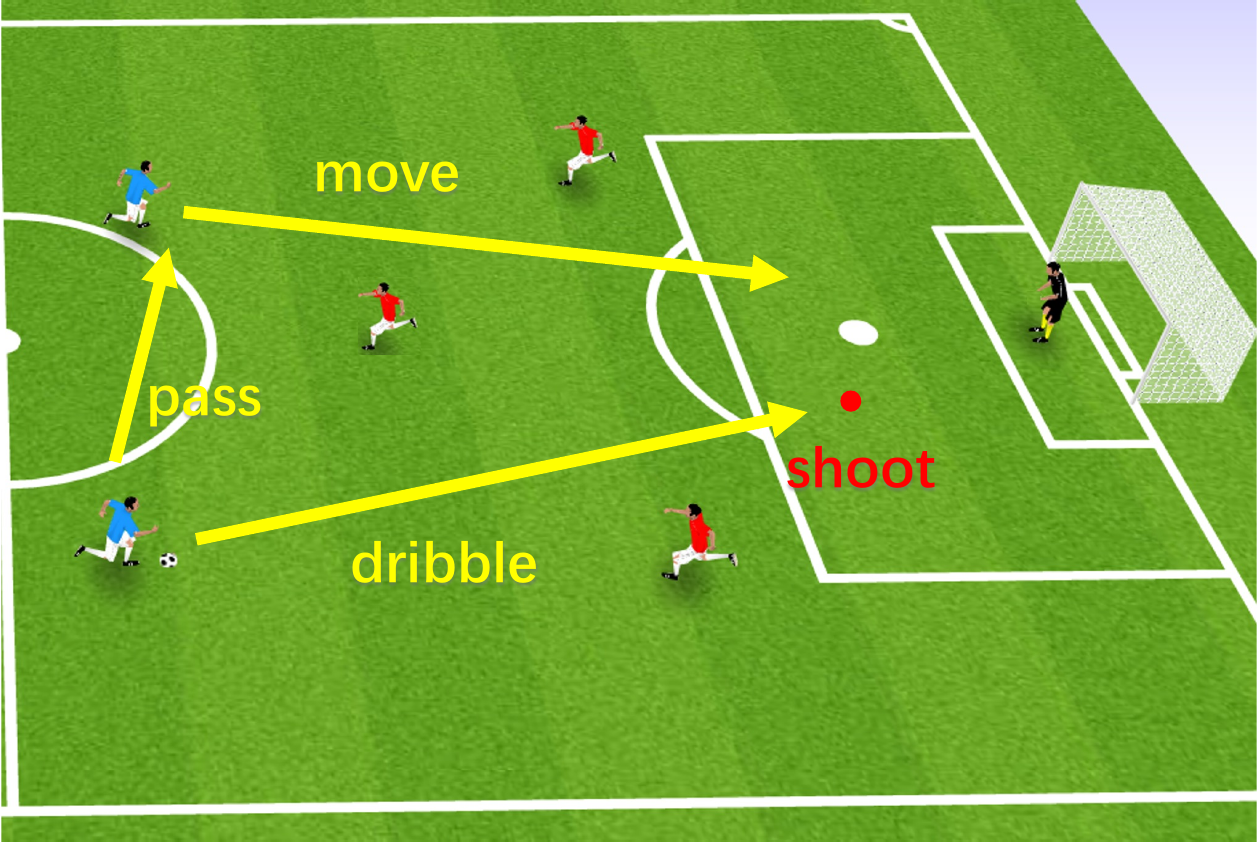}
}

\caption{We investigate decision-making with \textit{sparse action}, where agents must execute some actions sparsely due to restricted budgets or chances. Two examples of sparse action tasks are illustrated in the figures. \textbf{Left:} In the BattleCity game, the tanks fire sparingly due to restricted missile amounts. \textbf{Right:} In a football match, players repeatedly pass, dribble, and move before shooting.}
\label{fig:sparse_action_explanation}
\vspace{-0.7em}
\end{figure}

However, this assumption in classic RL methods is inconsistent with many real-world scenarios in which budgets or opportunities for executing specific actions are limited, i.e., decision-making under constrained resources~\citep{stock,air_combat,money_allocation,sparse_action_example}. For instance, in an air combat task, the pilot fires sparsely \citep{air_combat2} due to the limited missiles and tactical purposes. 
In this paper, we focus on decision-making tasks with limited budgets, including tasks like (1) Resource Management: Tasks where resources are finite and must be allocated judiciously (e.g., limited advertising budget); (2) Strategic Planning: Scenarios requiring optimal timing for action execution to maximize rewards (e.g., timing of football shot); (3) Tactical Games: Games where certain actions, like using a special ability in MOBA games, are restricted by cooldown periods.
For brevity, we name such actions as \textit{sparse action}, as shown in Fig. \ref{fig:sparse_action_explanation}. 
Certain action instances have elicited attention and have been researched independently \citep{stock,air_combat,football_score_eva,stock_review}. However, there has never been a method that generally solves all of these problems.

On the other hand, classic RL algorithms often explore all actions equally, such as exploration via $\epsilon$-greedy~\citep{nature_dqn,DBLP:conf/ccwc/DDQN,mnih2015human}, which executes actions randomly with probability, and max-entropy~\citep{SAC,HaarnojaTAL17}, which tries to increase the policy entropy. However, in sparse action tasks, such exploration mechanisms ignore the sparsity nature of these tasks. They might sample sparse actions early or excessively, making it difficult to acquire high-quality data. Consider air combat as an example: before engaging the opponent, the classic exploration mechanism, e.g., $\epsilon$-greedy, fires due to random action sampling. Such exploration on firing wastes the budgets of missiles and leads to ineffective sample collection. 
Some may concern that classic RL algorithm can already deal with sparse actions by giving negative reinforcements when a sparse action is executed frequently. We would like to clarify that these approaches is insufficient, as they do not differentiate between `frequent but necessary' actions and `frequent but unnecessary' actions.  We still require an algorithm that considers the discerns inherent sparsity of action and execute sparse actions without broadly penalizing action frequency.

This paper systematically studies the sparse action problem in RL. First, we define RL problems with sparse actions by introducing the Sparse-Action Markov Decision Process (SA-MDP). Compared to the standard Markov Decision Process, the SA-MDP considers the restriction on action execution number, i.e., in SA-MDP, some actions in the action space could only be executed for a limited number of times in every episode. In addition, we propose a policy optimization algorithm, Action Sparsity REgularization, \methodname. The motivation behind \methodname~is to explore each action with distinct preferences, which are measured by the action sparsity. \methodname~considers the action sparsity from two perspectives: (1) evaluating action sparsity by constraining action sampling during the exploration phase and (2) regularizing the policy learning using evaluated sparsity information during the training phase.
To be more specific, \methodname~selects one action when exploring the environment and constrains its sampling for several episodes. Action sparsity is evaluated based on the agent's performance when this action is constrained sampling. When training policy, \methodname~regularizes the policy learning with evaluated action sparsity, represented by action distribution.
We derive a regularized Bellman operator and prove its monotonicity and contractility. The algorithm finally converges to obtaining the regularized optimal value function by repeatedly applying the Bellman operator.

Our contributions are as follows: 
Firstly, we highlight the importance of decision-making under limited budgets. Such action has drawn much concern in previous works; while they study each sparse action separately, they still need a unified study over sparse action. 
In addition, we propose a generalized solution to such problem by formulating SA-MDP and presenting \methodname~method. In contrast to the classic exploration mechanism which treats all actions equally, we propose a novel idea that each action can be explored with different preferences. 
Our experiments show that the evaluated sparsity of different actions tends to vary. These intuitively sparse-executing actions (such as the shooting action in a football game) are truly evaluated to have a high sparsity. Moreover, except for tasks with explicit sparse action, we demonstrate that our approach outperforms the classic RL algorithm in common RL tasks, such as Atari.

\section{Related Work}
\label{sec:related_works}
This section first discusses prior research on sparse action tasks, followed by two areas closely related to our topic: constraining action sampling reinforcement learning \& regularization-based reinforcement learning. 

\subsection{Dealing with Sparse Action}
The existing literature extensively explores the execution of sparse actions in specific contexts \citep{sparse_action_example,sparse_action_example_2,sparse_action_example_3,sparse_action_example_4,sparse_action_example_5}. Notable examples of such scenarios encompass stock trading \citep{sparse_action_example_5,tnnls_pub1,tnnls_pub2}, football playing \citep{sparse_action_example,sparse_action_example_2}, air combat \citep{sparse_action_example_4}, and various decision-making situations involving limited resources \citep{limited_resources,limited_resources_2}, such as flyer distribution and monetary allocation \citep{money_allocation,tnnls_allocation}. These works study the property of a particular sparse action and present an action execution rule combined with a specific task.
Fire Control Model~\citep{air_combat} evaluates the critical aspect of firing a missile (e.g., missile flightpath, firing position) and develops specific firing criteria. 
Stock Market Decision~\citep{stock} explains how to purchase and sell stocks using historical stock data. IIDN \citep{money_allocation} designs the Serial Dictatorship Mechanism for resource allocation. In contrast, \methodname~abstracts the problem of sparse actions to a level that does not necessitate domain-specific data or heuristics. It is designed to be adaptable to a wide range of tasks by focusing on the underlying structure of decision-making in sparse action spaces. The proposed method dynamically adjust to the sparsity of actions without relying on external domain knowledge, making it inherently more flexible and applicable across diverse settings.

\subsection{Constraining Action Sampling RL}
Constraining action sampling is a method that enhances the exploration efficiency of RL algorithms by restricting agents from sampling specific actions \citep{tnnls_constrain}. This technique is crucial for tasks with sparse action spaces, where opportunities to execute particular actions are limited. Prior research has primarily focused on constraining action sampling by manually constructing prohibited state-action pairings. Action masking \citep{invalid_action_mask,action_mask_2} offers a knowledge-based pruning approach for limited action execution in RL, which masks the invalid actions. Nevertheless, action masking necessitates knowledge of the tasks. It may impede the performance of the policy owing to necessary action constraints. CSRL \citep{CSRL} incorporates prior domain knowledge as constraints/restrictions on the RL policy. They create numerous alternative policy constraints to preserve resilience in the face of the misspecification of individual constraints while capitalizing on advantageous constraints. An alternative method for constraining action sampling is to infer policy constraints from demonstrations \citep{guided_action_sampling1,guided_action_sampling2,constrain_sampling_action} and then constrain action sampling according to the policy constraints. MILP \citep{constrain_sampling_action_2} attempts to make decisions that depend only on a small number of simple state attributes. GPS~\citep{lev_gps} uses differential dynamic programming to provide appropriate guiding samples for policy search. In contrast to prior studies, which constrain sampling by needing explicitly established policy constraints or demonstrations, \methodname~constrains sampling by lowering the likelihood of action execution, as detailed in Sec. \ref{sec:method}.

\subsection{Regularization-based RL}
RL algorithms that utilize regularization as a core component have exhibited remarkable performance \citep{tnnls_pub3}. This section will focus on reward regularization, specifically regarding our proposed method.
A prominent instance of reward regularization is the Maximum Entropy regularization technique. This approach incorporates an entropy term to promote exploration, thereby fostering a more diverse and robust policy in the learning process~\citep{HaarnojaTAL17,SAC}. Specifically, max-entropy RL techniques \citep{max_entropy_1,max_entropy_2} also add a policy entropy term to the reward function and the origin reward from the environment. The rationale behind this method is to improve the randomness of the policy in order to explore additional states and actions. Recent research \citep{sparse_actor_critic,SDQN} demonstrates that Tsallis entropy regularization is also helpful in policy learning. RAC~\citep{RAC_regularized_actor_critic} illustrates that several traditional functions (such as the cosine/exp function) can be utilized as practical regularization functions in RL. Different from these works using various entropy terms or classic functions, the regularization in \methodname~is based on the evaluated action sparsity, which is described in detail in Sec. \ref{sec:method}.

\subsection{Sparse Reward Tasks}
\label{subsec:sparse_reward}
Sparse reward tasks are those where feedback (rewards) is infrequently given. This can make it difficult for an agent to understand which actions lead to positive outcomes \citep{sr_methods}. For instance, simple distance-to-goal reward shaping often fails in tasks where the agent must achieve some goal state, as it renders learning vulnerable to local optima1. A method has been introduced that uses an auxiliary distance-based reward based on pairs of rollouts to encourage diverse exploration \citep{sr_methods}. There are also previous works dealing with sparse reward issues under different settings \citep{sparse_reward_1,sparse_reward_2,sparse_reward_3,sparse_reward_4,sparse_reward_5,sparse_reward_6}. Among them, LOGO \citep{sparse_reward_1} and SQL \citep{sparse_reward_6} require a pre-collected offline dataset for policy training. In contrast, our method does not require any pre-collected data. MCAT \citep{sparse_reward_2} pertains to meta-reinforcement learning, which demands a set of similar tasks to learn a meta-policy, whereas our method focuses on training a policy for a single task. LOAD \citep{sparse_reward_3} is designed explicitly for object-interaction tasks with image-based observation input, requiring object segmentation techniques to highlight specific objects in the observation, such as the axe in a Minecraft game. HT-PSG \citep{sparse_reward_4} is tailored for continuous action spaces, while sparse action tasks are mainly for discrete action spaces. ExploRS \citep{sparse_reward_5} aligns with our setting, which trains policy under sparse reward setting without needing a set of tasks or dataset.

While both sparse action and sparse reward tasks deal with sparsity in RL, there are differences between them. Sparse reward tasks focus on the infrequency of feedback, while sparse action tasks deal with the decision-making under limited action budgets. Both present unique challenges in RL and require different strategies to address effectively. For a more comprehensive comparison, our experiments include a baseline \citep{sparse_reward_5} for handling sparse reward problem in Sec. \ref{sec:exp_sparse_action}.

\section{Problem Formulation}
In RL, the decision-making tasks are usually formulated as Markov Decision Process (MDP)~\citep{puterman2014markov, sutton2011reinforcement}, which can be described as a tuple $\gM = \left( \gS, \gA, P, r, \gamma, d_0  \right)$. Here $\gS$ is the state space. $\gA$ is the finite action space where $\gA = \{a_0, a_1, \cdots, a_{\vert \gA \vert - 1} \}$. $P$ denotes the transition probability and $r$ denotes the reward function. $\gamma$ is the discount factor that determines the weights of future rewards, and $d_0$ specifies the initial state distribution. A (stationary) policy $\pi: \gS \rightarrow \Delta (\gA)$ maps state space to the probability space over action space. The agent interacts with the environment as follows: at time step $t$, the agent observes a state $s_t$ from the environment and executes an action $a_t \sim \pi (\cdot |s_t)$, then the agent receives a reward $r(s_t, a_t)$ and transits to the next state $s_{t+1} \sim P(\cdot |s_t, a_t)$. 

The objective is to optimize a policy when specific actions in the action space are sparsely executed. Many factors might cause the sparse execution of actions, such as limited budgets or chances. In order to unify this class of problems, we call that these actions can only be executed for a limited number of times and propose \textit{Sparse-Action MDP} (SA-MDP) to formalize such a setting.

\begin{defn}[\textbf{Sparse-Action MDP}]
Given a MDP $\gM = \left( \gS, \gA, P, r, \gamma, d_0  \right)$, we define Sparse-Action MDP $\gM_{SA} = \left( \gS, \gA, P, r, \gamma, d_0, K  \right)$ by introducing action execution number limitation $K \in \mathbb{N}^+$. In Sparse-Action MDP, there is at least one action that could only be executed for at most $K$ times during one episode, i.e., $\exists a \in \gA$, such that for any trajectory $\tau, \sum_t \mathbb{I}(a_t^\tau=a) \leq K$, where $a_t^\tau$ denotes the action executed at timestep $t$ in trajectory $\tau$.
\end{defn}

Note that K reflects the inherent property of the task, which is unknown to \methodname. We aim to design an algorithm that adapts to different K values without requiring explicit adjustments.
Given a policy $\pi$, the value function is defined as the expected cumulative rewards it receives when starting with state $s$ and executing actions following $\pi$:
\begin{equation*}
\label{equation:v-function}
    V^\pi (s) = \expect \bigg[ \sum_{t=0}^{+\infty} \gamma^t r(s_t,a_t) \big| s_0 = s, a_t \sim \pi (\cdot|s_t)\bigg].
\end{equation*}
Correspondingly, the state-action value function is defined as the expected cumulative rewards, starting with state $s$, taking action $a$, and executing the following actions according to policy $\pi$.
\begin{equation}
\label{equation:q-function}
    Q^\pi (s, a) = r(s, a) + \gamma \expect_{s^\prime \sim P(\cdot|s, a)} \left[ V^{\pi} (s^\prime) \right].
\end{equation}
It is clear that an important connection between $V^{\pi} (s)$ and $Q^{\pi} (s, a)$ is shown as follows.
\begin{equation}
\label{equation:v-and-q}
        V^{\pi} (s) = \expect_{a \sim \pi (\cdot |s)} \left[ Q^{\pi} (s, a) \right].
\end{equation}
Then we define the optimal value function and state-action value function,
$
    V^*(s) = \max_{a\sim\pi} Q^{\pi} (s, a) \text{,}\; Q^* (s, a) = \max_{\pi} Q^{\pi} (s, a).
$
$V^* (s)$ and $Q^* (s, a)$ also hold the connection shown in Eq. (\ref{equation:q-function}) and Eq. (\ref{equation:v-and-q}). The optimal policy $\pi^*$ can be a deterministic policy which acts greedily with $Q^*$ (i.e., $\pi^* (s) = \argmax_{a} Q^* (s, a)$).

Given a policy $\pi \in \Delta (\gS)$, the associated Bellman operator $\gT^{\pi}$ is defined as, for any function $Q: \gS \times \gA \rightarrow \reals$,
\begin{equation*}
    \gT^{\pi} Q (s, a) = r(s, a) + \gamma \expect_{\substack{s^\prime \sim P (\cdot|s, a) \\ a^\prime \sim \pi (\cdot|s^\prime)}} \left[ Q (s^\prime, a^\prime)  \right]. 
\end{equation*}
From Eq. (\ref{equation:q-function}) and Eq. (\ref{equation:v-and-q}), we have that $Q^{\pi} (s, a)$ is the fixed point of $\gT^{\pi}$ (i.e., $\gT^{\pi} Q^{\pi} = Q^{\pi}$). We can define the Bellman optimality operator $\gT^{*}$ based on the Bellman operator. For any function $Q: \gS \times \gA \rightarrow \reals$,
$
    \gT^* Q = \max_{\pi} \gT^{\pi} Q.
$
It is known that $Q^*(s, a)$ is the fixed point of $\gT^{*}$ and $\gT^{*}$ is a $\gamma-$contraction with respect to the infinity norm (i.e., $\forall Q_1, Q_2 \in \gS \times \gA \rightarrow \reals, \; \| \gT^{*} Q_1 - \gT^{*} Q_2 \|_{\infty} \leq \gamma \| Q_1 - Q_2 \|_{\infty}$). With these properties, (approximate) value iteration repeatedly applies the Bellman optimality operator to attain the optimal state-action value function~\citep{DBLP:books/lib/BertsekasT96,error-bounds-for-avi}.

\section{Method}
\label{sec:method}
\begin{figure}[t]
\centering
\subfigure{
\includegraphics[width=0.98\linewidth]{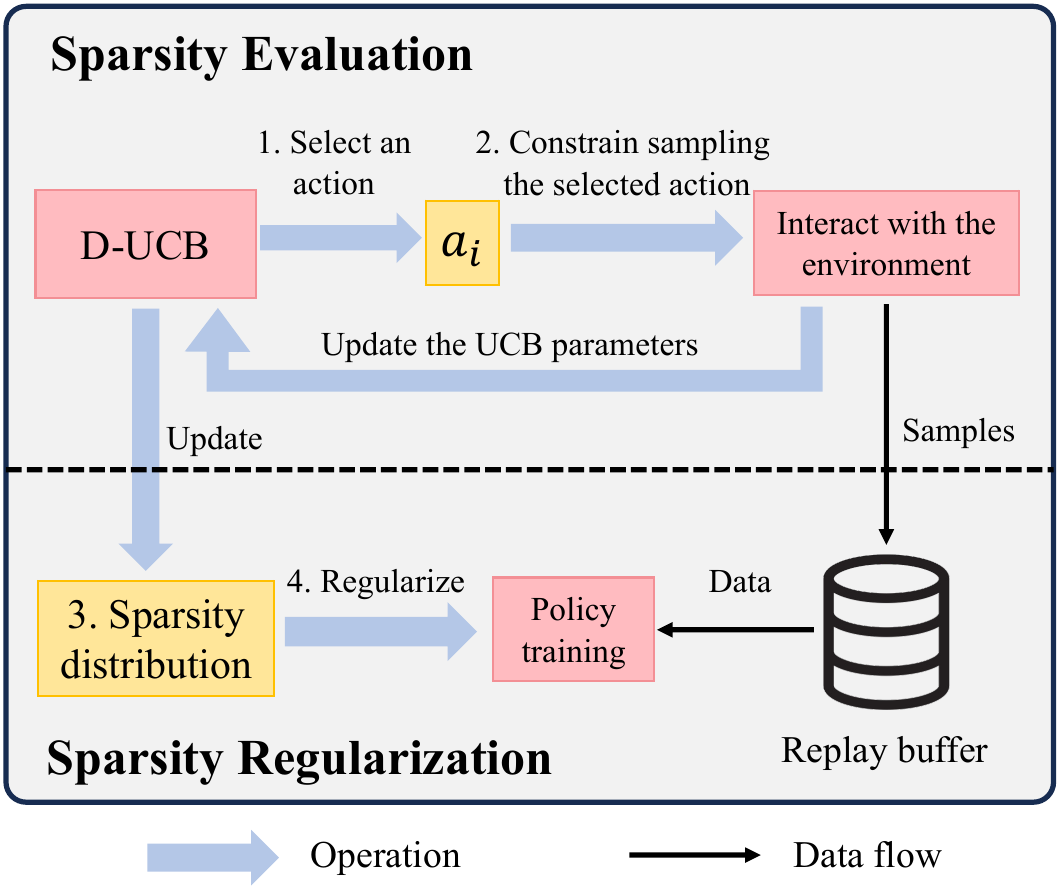}
}
\caption{Overall workflow of \methodname~comprises two parts: sparsity evaluation when exploring in the environment and sparsity regularization when training the policy.}
\label{fig:overall_framework}
\vspace{-0.5em}
\end{figure}

This section describes our approach to sparse action problems, which we refer to as \methodname. \methodname's central insight is to treat each action differently based on its sparsity. In the exploration phase, \methodname~specifically evaluates action sparsity via constrained action sampling. \methodname~regularizes policy learning with evaluated action sparsity during the training phase. Fig. \ref{fig:overall_framework} depicts the overall \methodname~workflow. Two key components of \methodname, namely sparsity evaluation via constraining sampling and policy optimization with sparsity regularization, will be discussed in the following two subsections, followed by a subsection that summarizes the entire training procedure of the \methodname~algorithm.

\subsection{Sparsity evaluation via constraining sampling}
\label{subsection_weights_update}

\methodname~is initially unaware of any information regarding action sparsity. Our central premise for evaluating action sparsity is that an action has a higher sparsity if executing it sparsely improves performance (i.e., higher reward). Note that accurately quantifying action sparsity is a complex challenge due to the dynamic and context complexity. Thus, \methodname~primarily attempts to offers a heuristic estimation by evaluating whether the overall performance of the agent improves when certain actions are executed less frequently. Based on this concept, \methodname~first selects an action and then observes the performance of the policy when constraint sampling the selected action. Here, two crucial issues arise: how to choose the action for evaluation and constrain action sampling.

\textbf{(Fig. \ref{fig:overall_framework}, 1) Select action to be evaluated.} It is not straightforward to select an action for evaluation, as random selection is inefficient because not all actions require constrained sampling. On the other hand, selecting the action based on the sparsity evaluated thus far may overlook other actions and result in a suboptimal choice. This problem can be viewed as the classical trade-off between exploration and exploitation, which naturally lends itself to modeling as a non-stationary bandit problem \citep{non_stationary_bandit_problem}. We treat each action as a bandit arm, and treat the discounted reward while constraining sampling as the bandit reward.
\methodname~uses the Discounted Upper Confidence Bounds (D-UCB) algorithm~\citep{discounted_UCB1,discounted_UCB2} to select action to be evaluated:
\begin{equation}
\label{eq:ucb_select}
    a_i = \arg\max_{a} \left( \mu(a) + c\sqrt{\frac{\log t(\tilde{\gamma})}{N_t(\tilde{\gamma},a)}}\right),
\end{equation}
where $\mu(a)$ is the estimated mean reward of the arm $a$. $c>0$, $N_t(\tilde{\gamma},a)$ denotes the discounted number of times $a$ is selected, $t(\tilde{\gamma})=\sum_{i=0}^{|\gA|-1}N_t(\tilde{\gamma},a_i)$ denotes the total discounted number of selections, and $\tilde{\gamma}$ is the discounted factor parameter of D-UCB.

\textbf{(Fig. \ref{fig:overall_framework}, 2) Constrain action sampling.} Constraining action sampling is an important technique in \methodname, with two effects: (1) reducing action sampling to avoid wasting budgets or opportunities to execute sparse actions, and (2) assisting in evaluating action sparsity. Given an action $a_i$ and a value function $Q(s,a)$, \methodname~constrains sampling $a_i$ via executing the behavior policy $\tilde{\pi}_{a_i}$:
\begin{equation}
\label{eq:sample_pi}
    \tilde{\pi}_{a_i}(a|s) \propto d_i(a) \exp \left( Q(s, a) \right),
\end{equation}
where $d_i (\cdot) = \left(\frac{1-\delta}{|\gA|-1}, \cdots, \delta, \cdots, \frac{1-\delta}{|\gA|-1}  \right)$ is an action distribution assigning a small probability $\delta < \frac{1}{|\gA|}$ to $a_i$ and an identical probability of $\frac{1-\delta}{|\gA|-1}$ to the other actions. The behavior policy can be viewed as a variant of Boltzmann policy \citep{boltzmann_policy}. It reduces the probability of executing $a_i$, thus constraining action sampling. After sampling with $\tilde{\pi}$ for $N$ episodes, statistical data of D-UCB, i.e., $\mu(a_i), t(\Tilde{\gamma}),N_t(\tilde{\gamma},a_i)$, get updated with the mean episodic reward of $N$ episodes. See Sec. 1.2 in \citep{discounted_UCB2} for more details about the update rule of D-UCB.

\textbf{(Fig. \ref{fig:overall_framework}, 3) Sparsity distribution.} Until now, action sparsity has been represented by $\mu(a)$: a larger $\mu(a)$ indicates higher action sparsity. \methodname~transforms $\mu(a)$ to \textit{sparsity distribution} $\piref(\cdot)$ using a Softmax function over the negative value of $\mu(a)$:
\begin{equation}
\label{eq:softmax_weight}
    \piref(a_i) \propto \exp(-\mu(a_i)).
\end{equation}

The sparsity distribution is a distribution over the action space. It intuitively assigns different probabilities to actions based on the evaluated action sparsity.

\vspace{-0.6em}
\subsection{Policy Optimization under Sparsity Regularization}
\label{subsection_our_method}

The previous section describes how \methodname~evaluates action sparsity and acquires sparsity distribution. This section describes how \methodname~employs sparsity distribution to enhance policy learning.

\textbf{(Fig. \ref{fig:overall_framework}, 4) Sparsity regularization.} \methodname~optimizes a regularized objective with an explicit constraint to the sparsity distribution:
\begin{equation}
\label{equation:objective}
    \max_{\pi} \expect \bigg[ \sum_{t=0}^{+\infty} \gamma^t (r (s_t, a_t) - 
    \lambda  \KL \left( \pi (\cdot|s_t) , \piref(\cdot) \right)) \big| d_0, \pi \bigg],
\end{equation}
where $\KL$ denotes Kullback-Leibler (KL) divergence. Different from the objective of SAC \citep{HaarnojaTAL17, SAC}, \methodname~uses a regularizes the policy with the sparsity distribution that reflects a distinct preference for each action. To optimize the objective, \methodname~searches for the regularized optimal policy, as shown in Proposition~\ref{prop:optimal_pi}.

Some may be concerned about the stability of APRE training with the sparsity regularization. 
At the early training stage, \methodname~explores randomly, similar to other RL algorithms. In this phase, all actions possess equal or close weights, and the regularization term is akin to maximum entropy RL \citep{max_entropy_1}. 
As training progresses, \methodname~employs the D-UCB algorithm for action selection and sparsity evaluation. The sparsity of action $a_i$ undergoes a soft update, as detailed in Sec. 1.2 of \citep{discounted_UCB1}, ensuring that the sparsity distribution stably changes. The D-UCB algorithm allows APRE to avoid actions that are either excessively frequent or infrequent, focusing instead on those with high sparsity potential. This makes \methodname~effectively evaluate all actions' sparsity.
The experimental results in Fig. \ref{expfig:weights_curves} further demonstrate that APRE can quickly evaluate action sparsity during the early training stage and maintain a stable sparsity distribution throughout the training process.

\begin{prop}[\textbf{Regularized Optimal Policy}]
\label{prop:optimal_pi}
    Given regularized optimal value function $Q_\Omega^* (s, a)$, the regularized optimal policy $\pi_{\Omega}^*$ can be obtained by
    \begin{equation}
    \label{equation:r-optimal-pi}
        \pi_{\Omega}^*(a|s) \propto \piref(a) \exp \left( \frac{Q^*_{\Omega} (s, a)}{\lambda} \right).
    \end{equation}
\end{prop}

We refer readers to Appendix \ref{appendix:proof1} for a comprehensive derivation of Proposition~\ref{prop:optimal_pi}. 
Eq. (\ref{equation:r-optimal-pi}) shows that the sparsity distribution acts directly on the probability distribution of the policy, which partially determines the policy distribution based on the evaluated action sparsity. 

One outstanding problem is obtaining regularized optimal value functions $ Q_\Omega^* (s, a)$. This is accomplished by \methodname's use of the regularized Bellman optimality operator proposed in Proposition \ref{prop:bellman_operator}. See Appendix \ref{appendix:proof2} for detailed proof.

\begin{prop}[\textbf{Regularized Bellman Optimality Operator}]
\label{prop:bellman_operator}

Under the regularized objective shown in Eq. (\ref{equation:objective}), the regularized Bellman optimality operator $\gT^*_{\Omega}$ is defined as, for any function $Q: \gS \times \gA \rightarrow \reals$,
\begin{equation*}
\label{equation:regularized-bellman-update}
\begin{split}
&\gT^*_{\Omega} Q(s, a) = r(s, a) +
    \\
    & \gamma \lambda \expect_{s^\prime \sim P(\cdot|s, a)} \left[  \log \left( \expect_{ a^\prime \sim \piref (\cdot|)} \left[ \exp \left( \frac{Q (s^\prime, a^\prime)}{\lambda}  \right)  \right] \right)   \right],
\end{split}
\end{equation*}
and then $Q^*_{\Omega}$ is the \textbf{fixed point} of $\gT^*_{\Omega}$. 
\end{prop}
\vspace{-0.5em}

\begin{algorithm}[t]
    \caption{Action Sparsity REgularization (\methodname)}
    \label{alg:method}
    \begin{algorithmic}[1]
    \State \textbf{Input:} episode number $N$ for evaluating sparsity, learning rate $\eta_\theta$, and Polyak update rate $\eta_{\rm pol}$. 
    \State Initialize Q-functions $Q_{\theta}$, target Q-functions $Q_{\Bar{\theta}}$, replay buffer $D$ and parameters of D-UCB.
        \While{task not done}
            \State // Sparsity evaluation
            \For{every $N$ episodes}
                \State Select an action to be evaluated with Eq. (\ref{eq:ucb_select}).
                \For{each environment step}
                    \State \parbox[t]{\dimexpr\linewidth-\algorithmicindent}{Interact with the environment with $\Tilde{\pi}$ \\
                    (Eq. (\ref{eq:sample_pi})) and store the transitions to $D$. \strut}

                    \State // Sparsity regularization
                    \State \parbox[t]{\dimexpr\linewidth-\algorithmicindent}{Sample batch data from $D$, update the \\ Q-function: $\theta \leftarrow \theta - \eta_\theta \hat{\nabla}_{\theta} J \left(\theta\right)$. \strut}
                    \State \parbox[t]{\dimexpr\linewidth-\algorithmicindent}{Update the target network parameters: \\ $\bar{\theta} \leftarrow \eta_{\rm pol} \theta+(1-\eta_{\rm pol}) \bar{\theta}$. \strut}
                \EndFor
                \State \parbox[t]{\dimexpr\linewidth-\algorithmicindent}{Update the statistical data of D-UCB and \\ the sparsity distribution $\piref$. \strut}
                  
            \EndFor
        \EndWhile
    \end{algorithmic}
\end{algorithm}

\subsection{Practical Algorithm}
\label{subsection_complete_apre}

This section summarizes the \methodname~practical training procedure as shown in Algorithm \ref{alg:method}. 
Overall, we optimize our regularized objective in an off-policy manner: The agent collects samples in the environments to update the Q-function parameterized with $\theta$. When exploring environment, \methodname~first selects an action $a_i$ using confidence bound (Eq. (\ref{eq:ucb_select})), and then collects samples following behavior policy (Eq. (\ref{eq:sample_pi})) that constrains sampling $a_i$ for $N$ episodes. The parameters of D-UCB are then updated with the mean episodic reward.

With samples collected from the environments, \methodname~repeatedly applies regularized Bellman optimality operator on the Q function $Q_{\theta}$ to obtain the optimal regularized state-action value function:
\begin{equation}
\begin{split}
 &Q^{k+1}_{\theta} (s, a) = r(s, a) +
    \\
    &\gamma \lambda \expect_{s^\prime \sim P(\cdot|s, a)} \left[  \log \left( \expect_{ a^\prime \sim \piref (\cdot)} \left[ \exp \left( \frac{Q^k_{\theta} (s^\prime, a^\prime)}{\lambda}  \right)  \right] \right)   \right].
\end{split}
\end{equation}
To complete the above update, \methodname~maintains a replay buffer $D = \{ (s_i, a_i, r_i, s_i^\prime) \}_{i=1}^m$. The target is to minimize the following empirical risk:
\begin{equation*}
\label{equation:q_loss}
    J (\theta) = \frac{1}{2m} \sum_{i=1}^m  \left( Q_{\theta} (s_i, a_i) - y \right)^2,
\end{equation*}
where $y = r_i + \gamma \lambda  \log \left( \expect_{ a^\prime \sim \piref (\cdot)} \left[ \exp \left( \frac{Q_{\Bar{\theta}} (s_i^\prime, a^\prime)}{\lambda}  \right)  \right] \right)$ and $\Bar{\theta}$ are the target parameters updated by the Polyak averaging \citep{polyak1992acceleration}. 

Finally, the regularized optimal policy can be directly obtained as analyzed in Eq. (\ref{equation:r-optimal-pi}):
\begin{equation*}
\label{equation:derive-pi}
    \pi (a|s) \propto \piref(a) \exp \left( \frac{Q_{\theta} (s, a)}{\lambda} \right).
\end{equation*}

\section{Theoretical Identification}

This section presents some theoretical properties of the proposed regularized Bellman operator, which justifies the feasibility of \methodname. For simplicity, we treat any $Q, \; \gT^*_{\Omega}Q \in \gS \times \gA \rightarrow \reals$ as a vector whose size is $\vert \gS \vert \cdot \vert \gA \vert$. We first show the monotonicity and contraction of $\gT^*_{\Omega}$.

\begin{prop}
\label{proposition:property}
The regularized Bellman optimality operator satisfies the following properties:

\begin{itemize}
    \item Monotonicity: for any $Q_1, Q_2 \in \gS \times \gA \rightarrow \reals$ such that $Q_1 \geq Q_2$,
        $\gT^*_{\Omega} Q_1 \geq \gT^*_{\Omega} Q_2$,
    where $ \geq $ means element-wise greater or equal.
    \item Contraction: for any $Q_1, Q_2 \in \gS \times \gA \rightarrow \reals$,
       $ \| \gT^*_{\Omega}Q_1 -  \gT^*_{\Omega} Q_2 \|_{\infty} \leq \gamma \| Q_1 - Q_2  \|_\infty$.
\end{itemize}

\end{prop}

See the Appendix for detailed proof. Proposition~\ref{proposition:property} indicates that the proposed regularized Bellman optimality operator is monotonic and contractional. Consequently, we can obtain the regularized optimal value function by repeatedly applying this operator. 

Like all regularization-based RL methods, adding the regularization term to the original objective modifies the original SA-MDP and the corresponding optimal value function. Such a disparity is quantified by Proposition~\ref{proposition:lower_bound}.

\begin{prop}[\textbf{Value Discrepancy}]
\label{proposition:lower_bound}
Let $\pi^*_{\Omega}$ denote the regularized optimal policy and $V^{\pi^*_{\Omega}}$ denote the value of $\pi^*_{\Omega}$ evaluated in the original SA-MDP. The prior probabilities on sparse action in each prior distribution are $\{\delta_1, \cdots, \delta_k\}$. Then we have that, for all $s \in \gS$ 
\begin{equation*}
   V^{\pi^*_{\Omega}} (s) \geq V^* (s) - \frac{C}{1-\gamma},
\end{equation*}
where $C = \lambda \max_a \log(1/\piref(a))$.
\end{prop}

See the Appendix \ref{appendix:proof4} for the proofs. 
Proposition \ref{proposition:lower_bound} demonstrates the effectiveness of \methodname~method, and ensures that the performance of the regularized policy remains within a reasonable range of the optimal value function. It indicates that $\lambda$ and $\piref$ govern the value discrepancy. This conclusion provides guidelines for hyper-parameter selection: setting a small $\lambda$ or inducing a small value of $\max_a \log(1/\piref(a))$ would make the regularized optimal policy maintain the optimality in the original SA-MDP. Note that the upper bound of the regularized value function $V^{\pi^*_{\Omega}}$ is inherently established by the nature of the optimal value function $V^*$.

\begin{figure*}[t]
\centering
\subfigure[Stock]{
\includegraphics[width=0.29\linewidth]{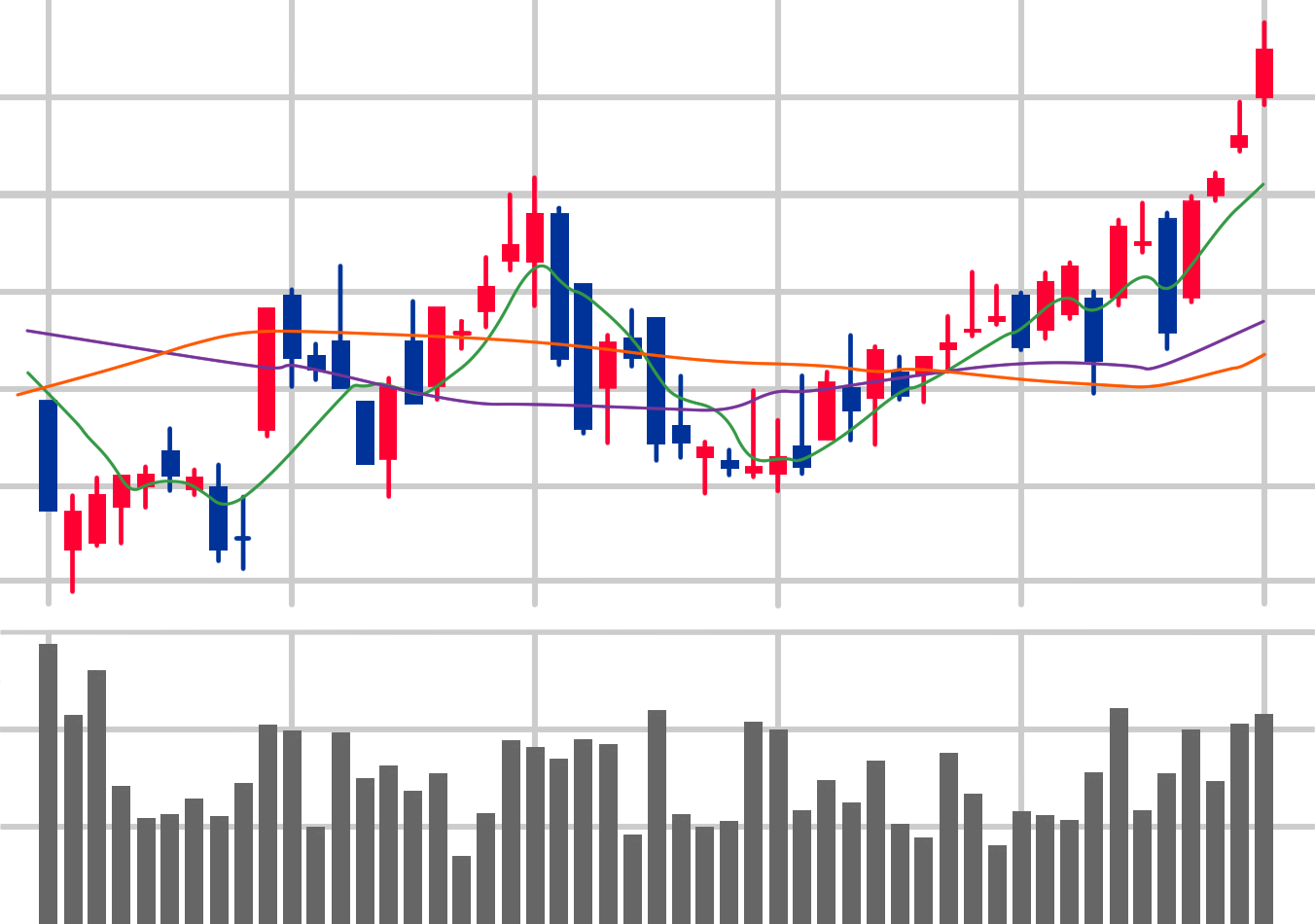}
\label{envfig_stock}
}
\centering
\subfigure[Gunplay]{
\includegraphics[width=0.29\linewidth]{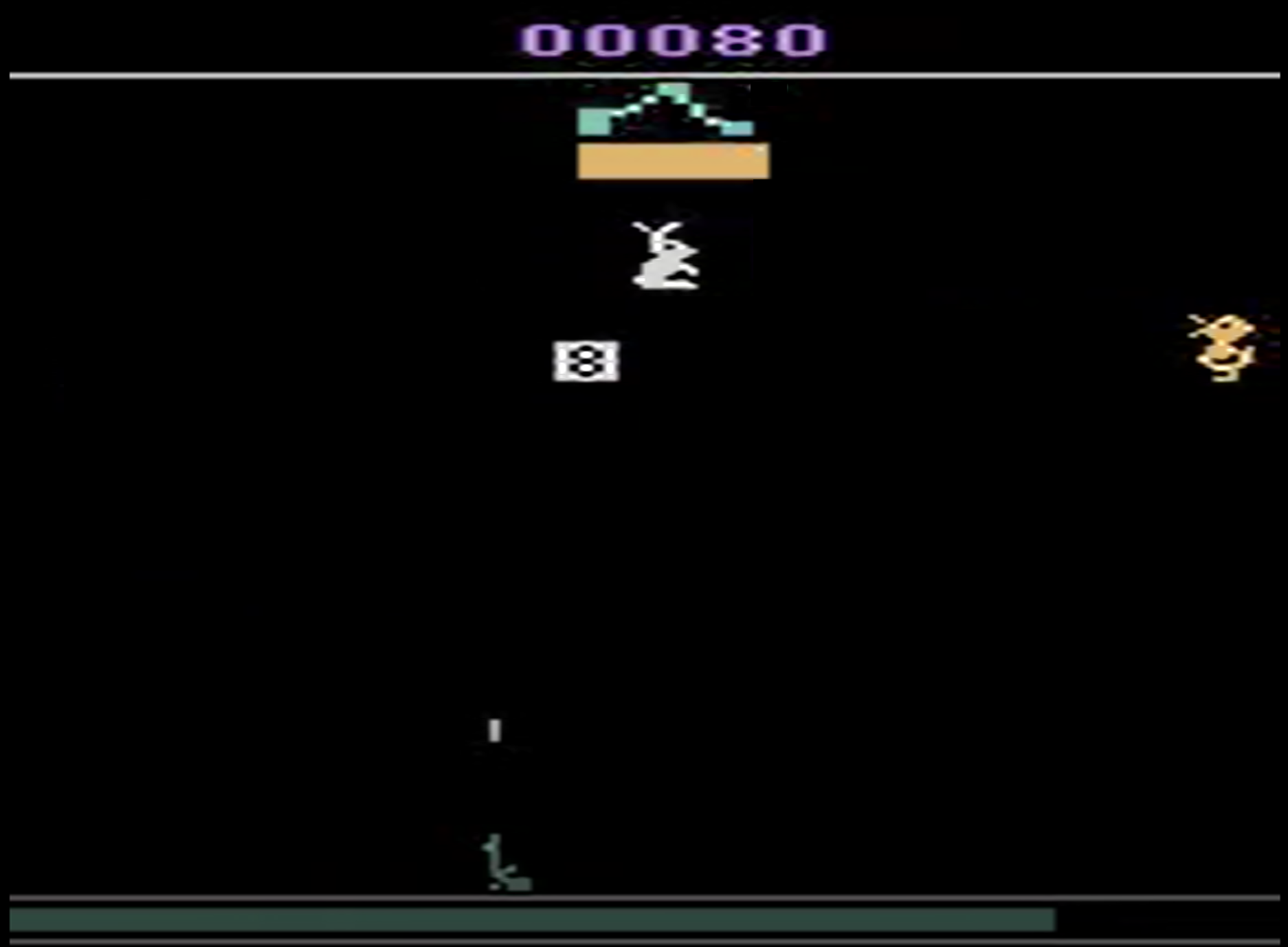}
\label{envfig_gun}
}
\centering
\subfigure[Football]{
\includegraphics[width=0.29\linewidth]{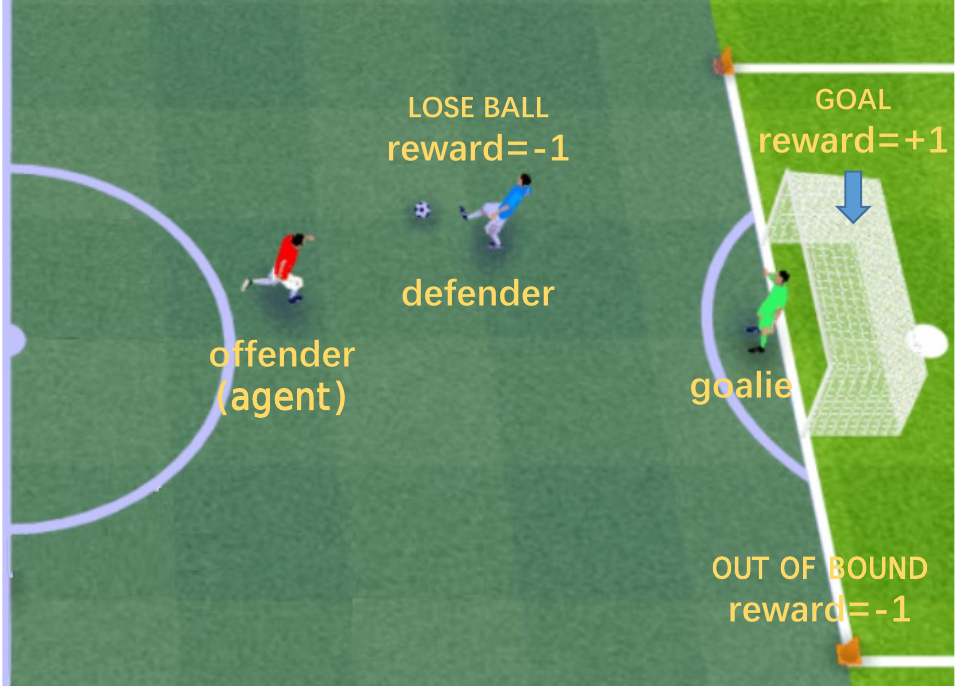}
\label{envfig_football}
}
\caption{Visualization of three sparse action tasks used in our experiments. (a) Stock: The agent buys/sells stock to earn profit. (b) Gunplay: The agent moves to the right and left while shooting at a moving target. (c) Football: The agent attempts to score while defended by a defender and goalkeeper.}
\label{exp_env}
\end{figure*}

\begin{figure*}[t!]
    \centering 
    \subfigure{
    \centering
    \includegraphics[width=0.98\textwidth]{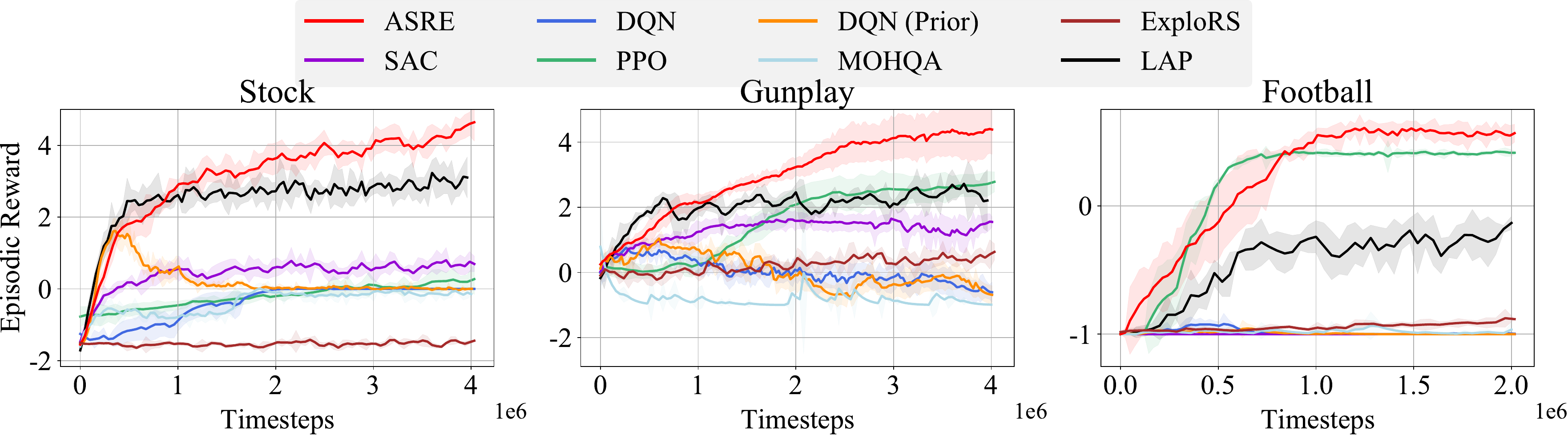}
    }
    \caption{Training curves of different policy optimization algorithms in diverse sparse action tasks. The x-axis represents the number of interaction steps, and the y-axis represents episodic reward. Shaded areas represent standard deviation across five runs.}
    \label{expfig:all_result} 
\vspace{-0.75em}
\end{figure*}

\section{Experiments}
\label{sec:experiments}

In this section, we empirically evaluate \methodname's performance through the following topics: (1) Main results on sparse action tasks (Sec. \ref{sec:exp_sparse_action}), (2) What happens during \methodname~training? (Sec. \ref{sec:exp_training_process} and \ref{subsection:ablation_study}), (3) How does \methodname~perform on different level of action sparsity (i.e., different values of K) (Sec. \ref{subsec:different_k}) and (4) Applicability of \methodname~on more general RL tasks (Sec. \ref{sec:exp_atari}).

\subsection{Main Results on Sparse Action Tasks}
\label{sec:exp_sparse_action}
We first introduce three sparse action tasks in our experiments, as shown in Fig. \ref{exp_env}. According to human knowledge, all these tasks involve sparse action, i.e., the buying/selling actions in Stock, the firing action in Gunplay, and the shooting action in Football.

\begin{itemize}[leftmargin=0.5cm]
    \item \textbf{Stock}: The stock trading task is constructed from real-world tick-level price and volume data on stocks in the Chinese A-share market \citep{sina_finance}, in which the agent earns a profit by buying at a low price and selling at a high price and can take one action of [\texttt{buy}, \texttt{sell}, \texttt{no-op}] at each decision step. In this task, \texttt{buy} and \texttt{sell} are considered as the sparse action, and the K constraint is 30. The stock code is ``XSHE: 000025'' (from January 10, 2017 to May 26, 2017).

    \item \textbf{Gunplay}: The agent scores by a shoot at the moving target. Hitting different targets yields different magnitudes of reward. This task is based on the Atari game ``Carnival-ram-v0'' in Gym benchmark~\citep{GYM}. In origin, the agent in the Carnival task is unlimited to shoot so long as task time is not used up, whereas it is a sparse action task when bullet capacity is limited. In the experiments, bullet capacity is limited to $5$ (i.e., the K constraint is $5$). The observation space is $\reals^{128}$. Available actions include [\texttt{no-op}, \texttt{fire}, \texttt{left}, \texttt{right}, \texttt{fire-left}, \texttt{fire-right}].

    \item \textbf{Football}: Football task is from HFO benchmark \citep{HFO}. The agent controls a football offender to score under the defense by a defender and a goalkeeper, controlled by the champion AI in 2012 RoboCup~\citep{agent2d_champlion}. The agent receives a reward of +1 when scoring and -1 when (1) losing possession of the ball, (2) running out of time, or (3) the ball is out of bounds. At each decision step, the agent may execute one of five high-level built-in actions: [\texttt{no-op}, \texttt{go to the ball}, \texttt{move}, \texttt{dribble}, \texttt{shoot}]. The \texttt{shoot} action is consider as the sparse action, and the K constraint is 4.
\end{itemize}

\begin{figure*}[t]
    \subfigure[Sparsity distribution during the training process]{
    \includegraphics[width=0.95\linewidth]{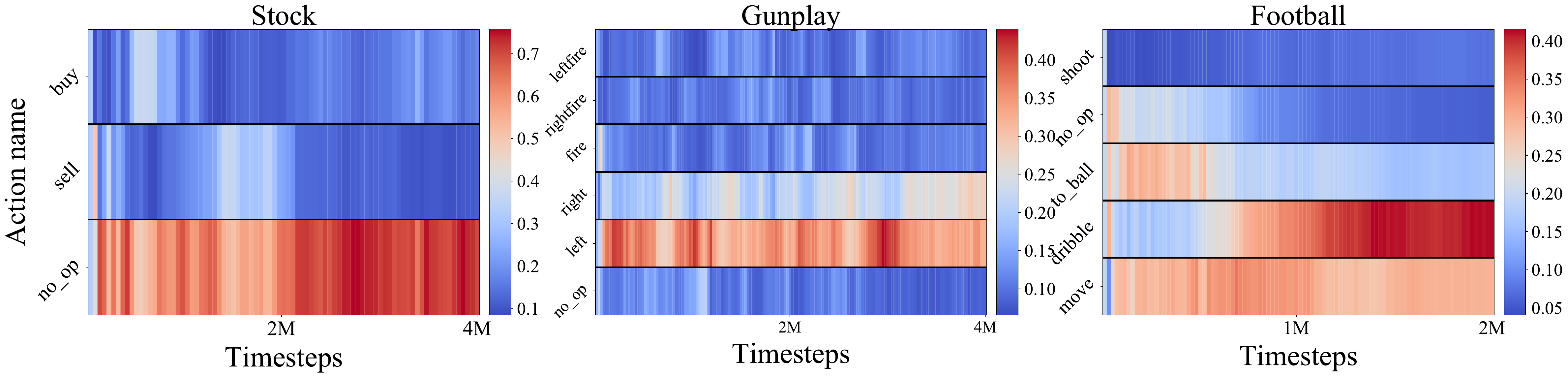}
    \label{expfig:weights_curves}
    }
    \subfigure[Frequency of executing sparse action.]{
    \includegraphics[width=0.69\linewidth]{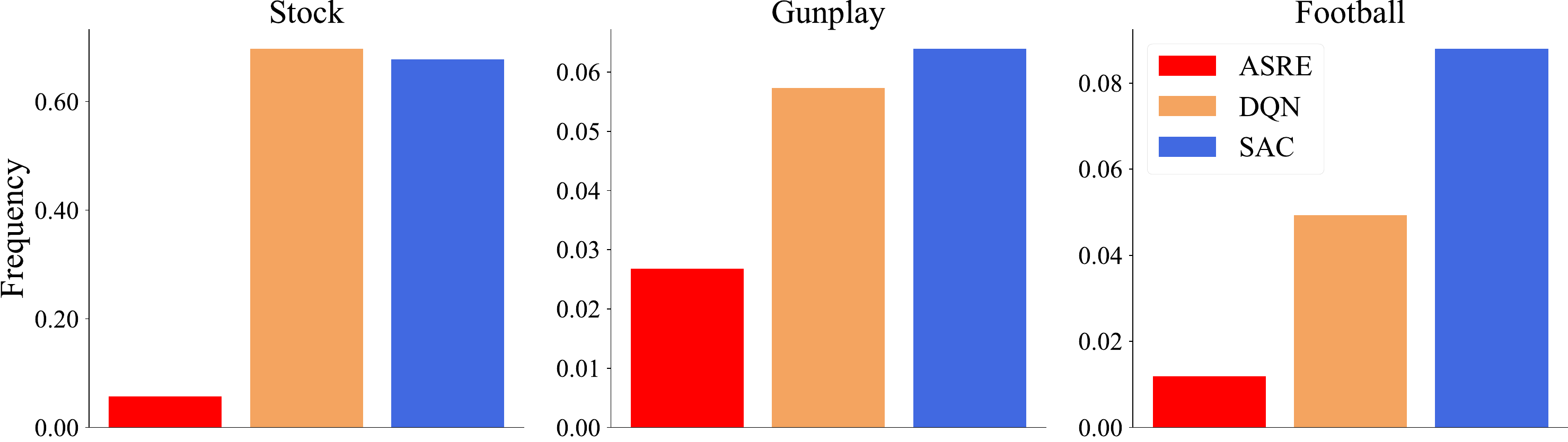}
    \label{exp_fig:freq}
    }
    \subfigure[Snapshot of shooting point.]{
    \includegraphics[width=0.26\linewidth]{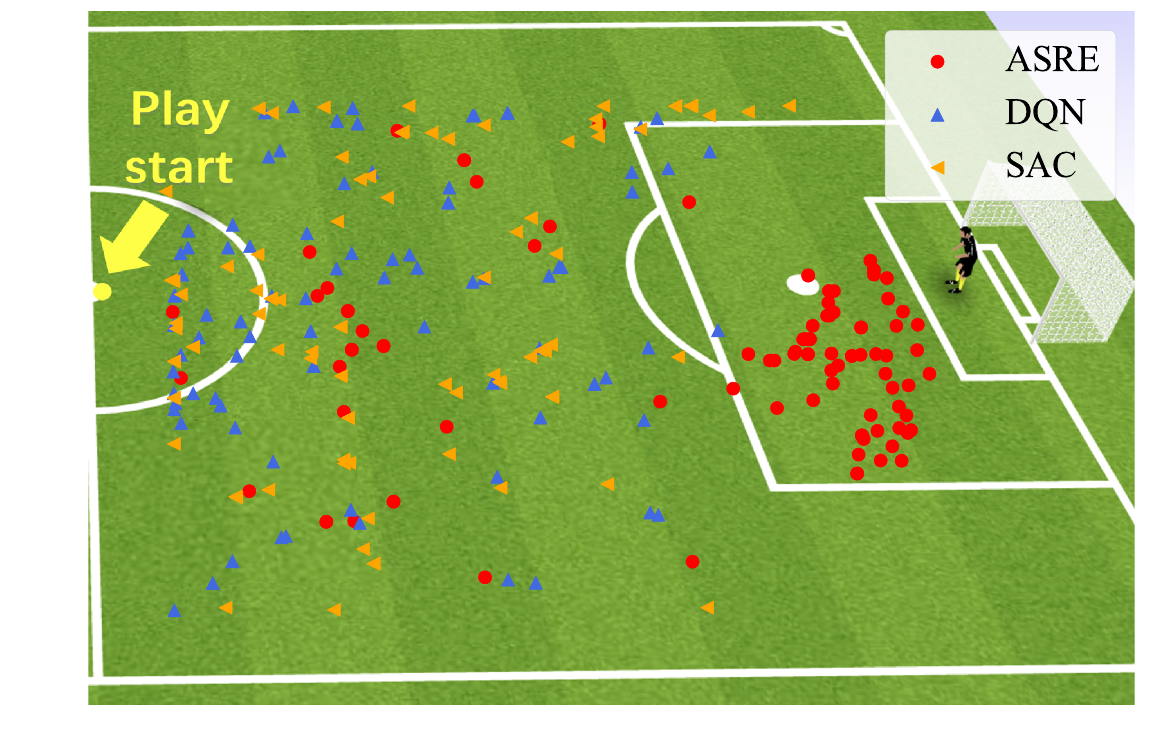}
    \label{expfig:shoot_dis_early}
    }
\caption{Sparsity evaluation and constraining action sampling during the training process. 
\textbf{(a):} Sparsity evaluation during the training process. Each row represents the probability of one action in sparsity distribution.
\textbf{(b):} Frequency of executing sparse action during the exploration stage. The frequency is calculated as (number of executing sparse actions) / (number of total decision steps).
\textbf{(c):} Snapshot of the shooting point of different agents.
}
\vspace{-0.75em}
\end{figure*}

We compare \methodname~with multiple RL baselines: (1) \textbf{DQN}~\citep{nature_dqn} and (2) \textbf{SAC} that explore with $\epsilon$-greedy and max-entropy mechanism, respectively. In our experiments, DQN employs various practical techniques \citep{DBLP:conf/aaai/rainbow}, including double Q, prioritized replay buffer, dueling network, and noisy layers. (3) \textbf{PPO}~\citep{PPO} explores using a stochastic policy and has been demonstrated to be effective in a variety of RL tasks \citep{PPO_implementation_matters}. (4) \textbf{DQN (Prior)} is a variant of DQN that penalizes the agent with a reward of 0.1 when executing these known sparse actions. DQN (Prior) can be viewed as a baseline with prior knowledge of action sparsity, but note that \methodname~is agnostic to any information about action sparsity.
(5) \textbf{ExploRS} \citep{sparse_reward_5} improves exploration by learning an intrinsic reward function in combination with exploration-based bonuses to maximize the agent's utility w.r.t. extrinsic rewards.
(6) \textbf{MOHQA} \citep{mohqa} proposes a novel neural architecture called Modulated Hebbian plus Q network architecture. The key idea is to use a Hebbian network with rarely correlated bio-inspired neural traces to bridge temporal delays between actions and rewards when confounding observations result in inaccurate TD errors.
(7) \textbf{LAP} \citep{lap_pal} is an improvement over prioritized experience replay (PER). It simplifies PER by removing importance sampling weights and setting the minimum priority to 1, reducing bias. At the same time, it uses a prioritized approximation loss, which uniformly sampled loss equivalent of LAP.

Fig. \ref{expfig:all_result} illustrates the training progress of various reinforcement learning algorithms on sparse action tasks. 
Overall, \methodname~agent can effectively complete three sparse action tasks and outperform all baselines. We observe that \methodname~learns faster in Stock and Gunplay, where sparse actions occupy a large portion of the action space, and achieves a higher final score in Football. DQN exhibits limited progress in the Gunplay and Football tasks and converges to a low score in the Stock task. This can be attributed to the $epsilon$-greedy exploration strategy, which indiscriminately explores all actions and inefficiently allocates resources for executing sparse actions, rendering it unsuitable for sparse action tasks. Although SAC outperforms DQN, it also plateaus at a low reward level, suggesting that max-entropy exploration struggles to learn the execution of sparse actions efficiently. In the Football task, PPO achieves performance comparable to \methodname. However, in the Stock and Gunplay tasks, PPO's scores are significantly lower than those of \methodname. PPO outperforms DQN and SAC in the Gunplay and Football tasks, indicating that stochastic exploration is more effective in sparse action tasks than max-entropy and $\epsilon$-greedy exploration strategies.
DQN (Prior) yields low scores across all three tasks, signifying that merely incorporating a negative reward is insufficient for effective decision-making in sparse action tasks. ExploRS exhibits minimal improvement in all tasks, which can be ascribed to its primary focus on addressing sparse reward tasks rather than sparse action tasks. These results indicate that the techniques used in the sparse reward domain are ineffective in solving sparse action tasks. 
MOHQA method focuses on addressing the inaccurate TD errors, thus improving the performance over the vanilla DQN algorithm. 
Meanwhile, the LAP algorithm, by focusing on the efficient utilization of the replay buffer, outperforms or is comparable to other baselines. This performance differential may be ascribed to our sample selection strategy, which effectively mitigates value errors in sparse action scenarios. Despite this, LAP does not quite reach the performance of \methodname.

\vspace{-0.8em}
\subsection{Training Process of \methodname}
\label{sec:exp_training_process}

The preceding subparagraph suggests that \methodname~can effectively train policy in sparse action tasks. In this subsection, we study how \methodname~works by demonstrating the sparsity evaluation and constraining action sampling.

\textbf{Sparsity evaluation}. Fig. \ref{expfig:weights_curves} depicts the sparsity distribution evaluated during training for three sparse action tasks. Lower action probability (cold tone in the figure) reflects greater action sparsity in sparsity distribution. Overall, the action sparsity evaluated is consistent with human intuition regarding tasks. For instance, buy/sell actions in Stock are considered high sparsity, as frequently buying/selling induces high transaction fees. \methodname~can rapidly evaluate the action sparsity in the very early training stage. The evaluated sparsity distribution is nearly stable throughout the training process. In conjunction with the results of the previous section's experiments, the minute change in sparsity distribution does not hinder policy learning. 
In Gunplay, three actions involving shooting and no-op are evaluated as low probability. In contrast, actions involving moving right/left are preferred. 
In football, \methodname~suggests that shoot and no-op are in high sparsity, while the sparsity of dribble action is lowest.

\begin{figure*}[t!]
    \centering  
    \includegraphics[width=0.95\textwidth]{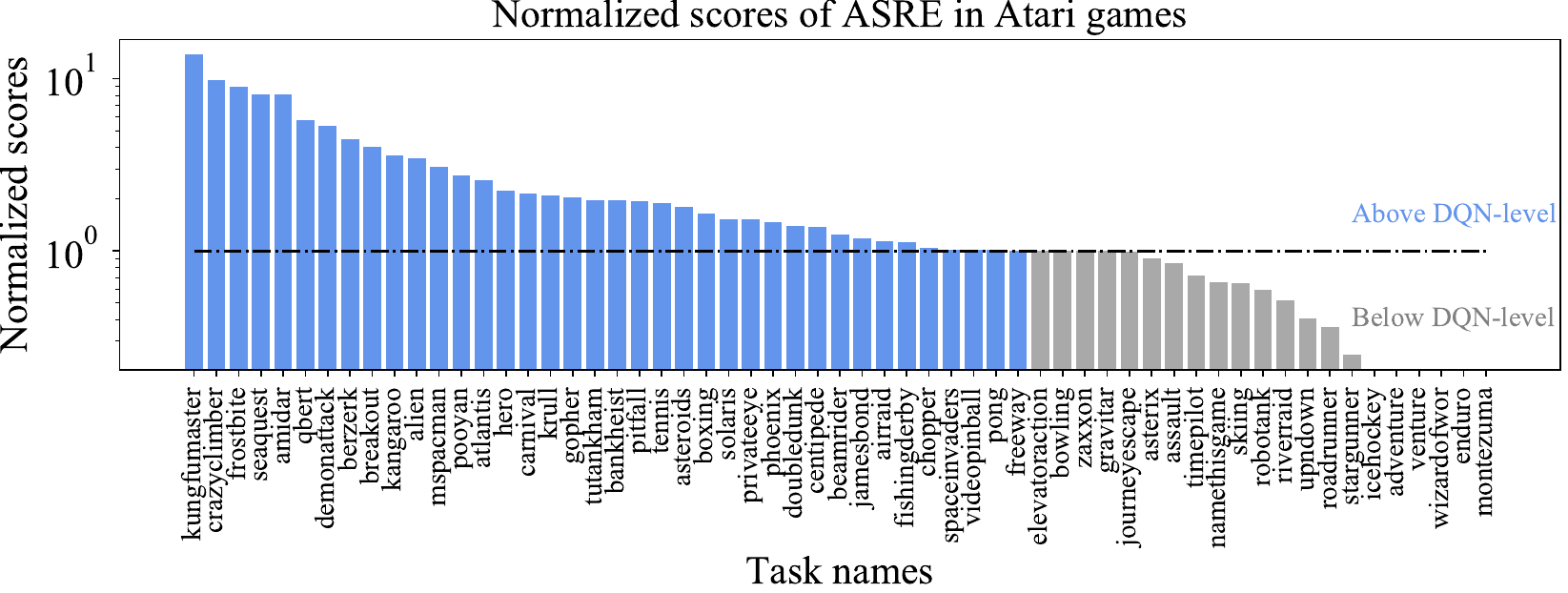}
    \centering
    \caption{Comparison of \methodname~with DQN in 59 Atari games. The normalized scores are calculated as (\methodname~score - random score) / (DQN score - random score). \methodname~outperforms DQN on 38 of the evaluation Atari games.} 
    \label{expfig:atari} 
\vspace{-0.5em}
    
\end{figure*}

\textbf{Constraining action sampling}. The objective of \methodname~is to evaluate action sparsity via constrained action sampling. Fig. \ref{exp_fig:freq} shows the frequency of sparse action execution by various algorithms throughout the entire exploration phase. Contrary to SAC and DQN, the constraining action sampling technique in \methodname~can reduce the frequency of executing sparse action. SAC and DQN agents execute buying/selling actions at greater than $60\%$ of decision steps, a high-frequency trading operation in the Stock task. \methodname~agent only trades at less than $10\%$ of decision steps, correspondingly. This result demonstrates that \methodname~effectively addresses the issue of sampling sparse action excessively that plagues traditional RL methods. Besides, we record the shooting point of \methodname, DQN, and SAC agents while exploring the Football task at 200k training timesteps, shown in Fig. \ref{expfig:shoot_dis_early}. When interacting with the environment, DQN and SAC agents consume shooting opportunities when they are far from the objective, making it more difficult to collect high-quality samples. Contrary to the DQN/SAC agents, the \methodname~agent can take the ball closer to the goal and shoot from more advantageous areas due to more careful sparse action execution caused by constraining action sampling.

\begin{figure}[htbp]
    \centering  
    \includegraphics[width=0.4\textwidth]{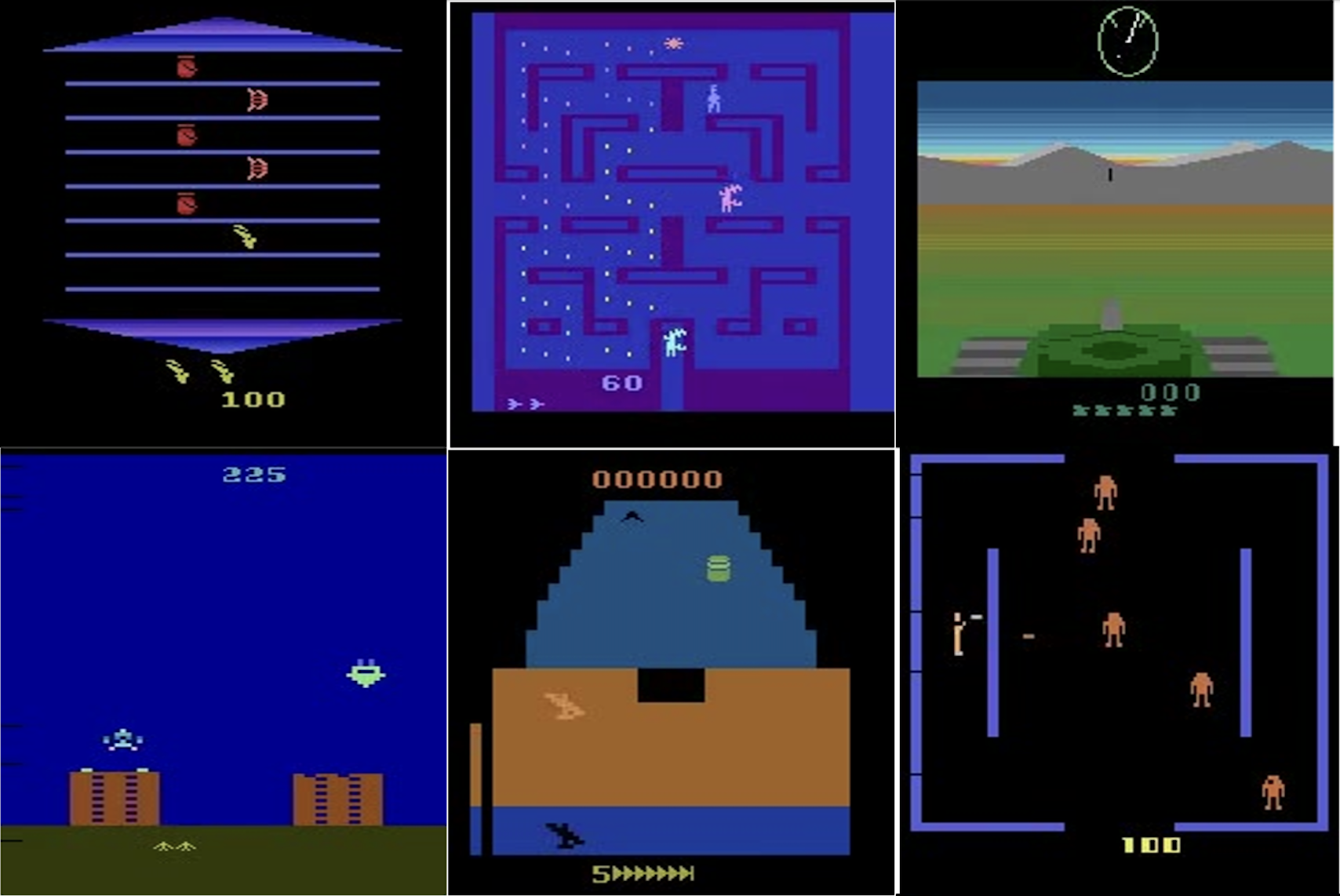}
    \centering
    \caption{Snapshots of the Atari games.} 
    \label{expfig:atari_game} 
\vspace{-0.3em}
\end{figure}

\subsection{Performance Under Different Levels of Sparsity}
\label{subsec:different_k}

\begin{figure}[h]
\centering
\subfigure{
\includegraphics[width=0.98\linewidth]{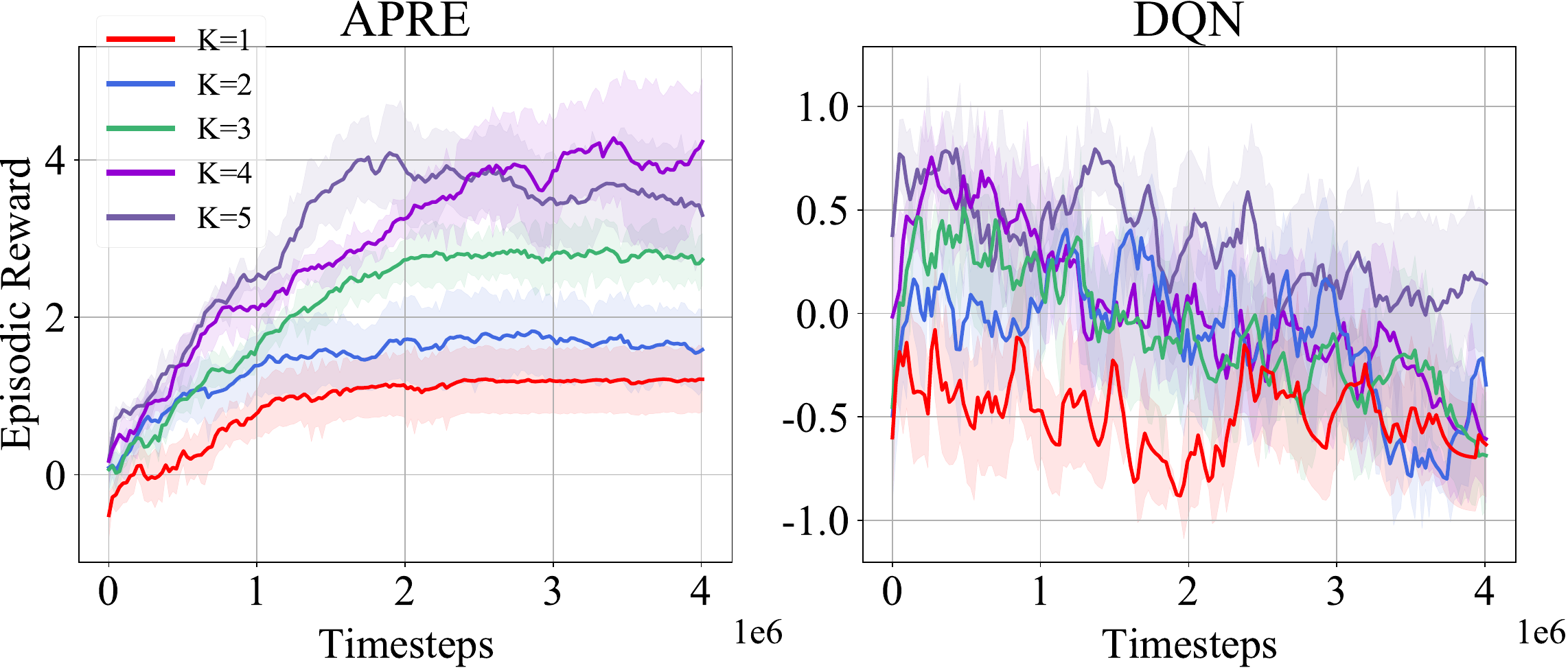}
}
\caption{Training curves with different K.}
\label{fig:diff_k}
\vspace{-0.5em}
\end{figure}

We further examine the impact of K in the SA-MDP framework on algorithm performance. Specifically, we vary the bullet count (i.e., the maximum number of shots) within the set {1, 2, 3, 4, 5} for Gunplay tasks. Fig. \ref{fig:diff_k} illustrates the training curves of \methodname~and DQN for different K values. We observe that our method enhances policy performance across various K values. A larger K results in superior policy performance, as it corresponds to an increased number of shots, enabling the agent to achieve a higher score. When K is sufficiently large (i.e., $K\geq4$), the convergence performance of \methodname~approaches a similar level. In contrast, DQN exhibits difficulty in enhancing policy performance across all five K values.

\subsection{Applicability on More General Tasks}
\label{sec:exp_atari}

To further validate the applicability and robustness of the \methodname~method across various decision-making tasks, we also include experiments in 59 Atari games. \textit{Atari} is an RL benchmark that provides a variety of games, such as shooting, expedition, fighting, car racing, and more, as shown in Fig. \ref{expfig:atari_game}.
We compare \methodname~with DQN (with four implementation techniques \citep{DBLP:conf/aaai/rainbow}: double Q, prioritized replay buffer, dueling network, and noisy layers) in 59 Atari games on the Gym platform, with vector observation of $\mathbb{R}^{128}$, and action space of $\mathbb{R}^{6}$ or $\mathbb{R}^{18}$. The agent is trained for 4M timesteps and then evaluated for 100 episodes in each experiment. The normalized \methodname~over DQN scores in 59 Atari games are displayed in Fig. \ref{expfig:atari}. In 38 evaluation tasks, \methodname~outperforms DQN. The results demonstrate that action sparsity is a significant factor in enhancing the learning algorithm's performance and that, in addition to RL tasks with sparse actions, \methodname~can also be applied to more general RL tasks. This underscores the method’s versatility and potential for generalization beyond tasks with clearly defined sparse actions.

\subsection{Ablation Study}
\label{subsection:ablation_study}

\begin{figure}[h!]
    \centering 
    \includegraphics[width=0.48\textwidth]{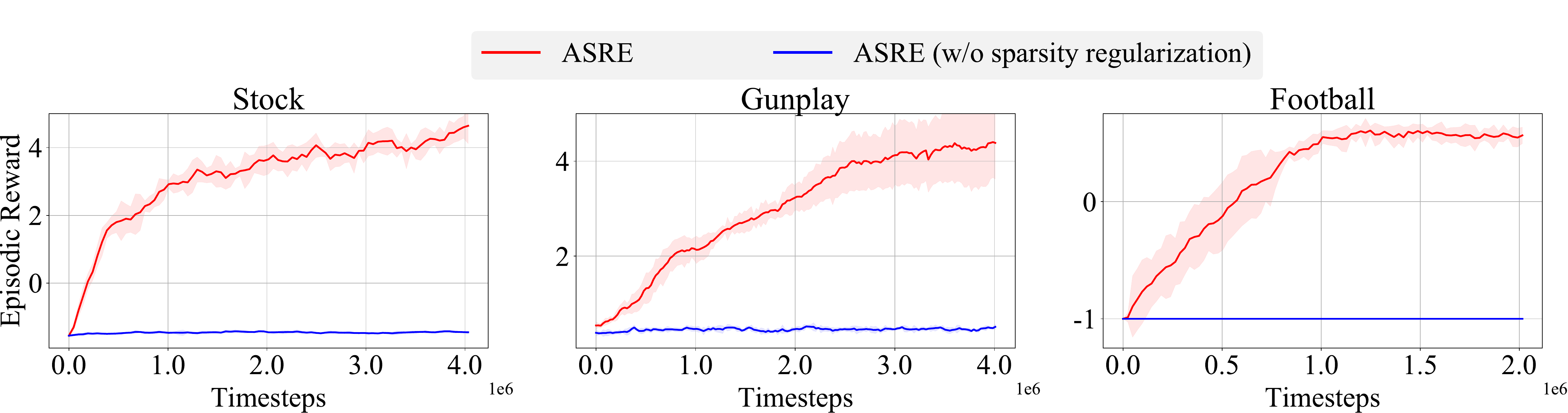}
    \centering
    \caption{Ablation study of sparsity regularization. The blue curve denotes the training curve of \methodname~without sparsity regularization.}
    \label{expfig:sparsity_regularization} 
\end{figure}

\textbf{Sparsity regularization}. 
In order to assess the efficiency of sparsity regularization, we conduct an ablation study. 
The training curve for \methodname~with and without sparsity regularization (i.e., updating Q function and deriving optimal policy without sparsity distribution regularization) is shown in Fig. \ref{expfig:sparsity_regularization}. It is clear from our observations that \methodname~does not learn much without sparsity regularization, and the training curve almost remains horizontal. This result highlights the significance of sparsity regularization in \methodname.

\begin{table}[htbp]
    
    \centering

    \caption{Comparison of different magnitudes of regularization coefficient ($\lambda$). Each data is averaged over the episodic reward of the last $5$ checkpoints. The number after $\pm$ is the standard deviation over $5$ random seeds.}
    \begin{tabular}{c|c|c|c}

    \toprule
    \diagbox{$\lambda$}{Task} & Stock & Gunplay & Football \\   
    \midrule
    0.005 & $4.28 \pm 0.41$ & $3.24 \pm 0.36$ & $0.50 \pm 0.08$  \\
    \midrule
    0.01 & $\textbf{4.72} \pm 0.37$ & $\textbf{4.41} \pm 0.79$ & $\textbf{0.57} \pm 0.07$  \\
    \midrule
    0.05 & $2.75 \pm 0.39$ & $2.46 \pm 0.15$ & $-0.09 \pm 0.06$  \\
    \midrule
    0.2 & $-0.04 \pm 0.23$ & $0.46 \pm 0.08$ & $-0.84 \pm 0.04$  \\
    \bottomrule
    \end{tabular}
    \label{exptab:different_scale}
\end{table}

\textbf{Regularization coefficient $\lambda$.} The final score of the \methodname~agent trained with various regularization coefficient magnitudes $\lambda$ is shown in Tab. \ref{exptab:different_scale}. We observe that \methodname~works well when $\lambda$ is low. In general, $\lambda=0.01$ is a good parameter for all environments, and larger $\lambda$ will result in a reduction in performance. 
This is because a large $\lambda$ causes the policy learning to be overly dependent on the sparsity distribution and expose it to bad policy. When $\lambda$ is getting smaller (see $\lambda=0.005$), the policy learning depends less on the evaluated action sparsity, and there is a slight decline in the agent's performance.

\section{Conclusion \& Limitation}

\label{sec:conclusion}
This paper highlights sparse-executing action in RL. Classical RL algorithms are unaware of the sparsity nature of sparse action and are unable to explore and learn efficiently. To solve this problem, we put forth SA-MDP to formalize sparse action RL tasks and propose \methodname~algorithm, which evaluates action sparsity via constraining action sampling and then uses the evaluated action sparsity to train policy.
Experiments on various RL tasks present that \methodname~can efficiently evaluate action sparsity and constrain action sampling. Moreover, we demonstrate how our method can effectively optimize policy in both Atari games and sparse action tasks.

To our knowledge, this is the first work to systematically study sparse-executing actions in RL. While \methodname~can effectively complete RL tasks with sparse action, there are still some limitations. 
The first limitation regards the training stability. The sparsity distribution, which directly affects policy learning, changes every $N$ episodes, which may be a volatile factor for the training process. Besides, \methodname~repeatedly selects action and constrains action sampling, resulting in that the optimized policy is significantly different from the exploration policy. A future direction to improve \methodname~is to resolve the problem of off-policy sample collection by utilizing current policy to interact with the environment on some of the exploration episodes. 
The second limitation regards the application scope: \methodname~is initially designed for discrete action as it is infeasible to evaluate the sparsity of continuous actions. As much as possible, we would like to extend \methodname~to cases with continuous action space, e.g., by utilizing a neural network to approximate the sparsity distribution.
We hope that future work can explore these interesting questions and make a step further forward for sparse action tasks.

\appendix

\subsection{Omitted Proofs}
\label{appendix:all_proof}

\subsubsection{Proof of Proposition \ref{prop:optimal_pi}}
\label{appendix:proof1}
\begin{proof}
Under the regularized objective defined by Eq. (\ref{equation:objective}), the regularized value functions are defined as:
\begin{equation}
\label{eq:regularized_value_func}
    Q^\pi_\Omega(s,a)=r(s,a)+\gamma \expect_{s'\sim P(\cdot|s,a)}[V^\pi_{\Omega}(s')],
\end{equation}
\begin{equation*}
\begin{split}
    V^\pi_\Omega(s)=\expect_{a\sim \pi(\cdot|s)}[Q^\pi_\Omega(s,a)] 
    -\lambda  \KL \left( \pi (\cdot|s) , \piref(\cdot) \right),
\end{split}
\end{equation*}

The regularized optimal value functions and the regularized optimal policy are $V^*_{\Omega} (s) = \max_{\pi} V^{\pi}_{\Omega} (s) \text{,}\; Q^*_{\Omega} (s, a) = \max_{\pi} Q^{\pi}_{\Omega} (s, a)$, and $\pi_{\Omega}^*=\argmax_{\pi}V^\pi_\Omega(s)$, respectively. $V^*_{\Omega}$ and $Q^*_{\Omega}$ also hold the connection in Eq. (\ref{eq:regularized_value_func})~\citep{RMDP}.

We can obtain that $\forall s \in \gS$,
\begin{equation*}
\begin{split}
    \pi^*_{\Omega} (\cdot|s) = \argmax_{\pi} \bigg\{ \expect_{a \sim \pi (\cdot|s)} \left[ Q^*_{\Omega} (s, a) \right] - 
    \\
    \lambda \KL \left( \pi (\cdot|s), \piref (\cdot) \right) \bigg\}.
\end{split}
\end{equation*}
Differentiate the right side with respect to $\pi (a|s)$ and let the derivative equal zero; we obtain that
\begin{equation*}
    \sum_{a \in \gA} Q^{*}_{\Omega} (s, a) - \lambda \log \left( \frac{\pi^* (a|s)}{\piref (a)} \right) - \lambda = 0.
\end{equation*}
Solving this formula yields
\begin{equation*}
    \pi^*_{\Omega} (a|s) \propto \piref(a) \exp \left( \frac{Q^*_{\Omega} (s, a)}{\lambda} \right).
\end{equation*}
\end{proof}

\subsubsection{Proof of Proposition \ref{prop:bellman_operator}}
\label{appendix:proof2}
\begin{proof}
In Appendix \ref{appendix:proof1}, we show that
\begin{equation*}
\begin{split}
    &Q^*_{\Omega} (s, a) = r(s, a) + \gamma \expect_{s^\prime \sim P(\cdot|s, a)} \left[ V^{*}_{\Omega} (s^\prime) \right], \text{with}
    \\
    & V^{*}_{\Omega} (s) = \expect_{a \sim \pi^*_{\Omega} (\cdot |s)} \left[ Q^{*}_{\Omega} (s, a) \right] - \lambda \KL \left(\pi^*_{\Omega}(\cdot|s), \piref(\cdot)  \right).
\end{split}
\end{equation*}
Substitute $\pi^*_{\Omega} (a|s)$ with $\piref(a) \exp \left( \frac{Q^*_{\Omega} (s, a)}{\lambda} \right)$ and we obtain that
\begin{equation*}
    V^{*}_{\Omega} (s) = \lambda \log \left( \expect_{ a \sim \piref (\cdot)} \left[ \exp \left( \frac{Q^*_{\Omega} (s, a)}{\lambda}  \right)  \right] \right).
\end{equation*}
Then we have that
\begin{equation*}
    \begin{split}
        &Q^*_{\Omega} (s, a) = r(s, a) + 
        \\
        & \gamma \lambda \expect_{s^\prime \sim P(\cdot|s, a)} \left[  \log \left( \expect_{ a^\prime \sim \piref (\cdot)} \left[ \exp \left( \frac{Q^*_{\Omega} (s^\prime, a^\prime)}{\lambda}  \right)  \right] \right)   \right],
    \end{split}
\end{equation*}
which shows that $Q^*_{\Omega}$ is the fixed point of $\gT^*_{\Omega}$.

\end{proof}

\subsubsection{Proof of Proposition \ref{proposition:property}}
\label{appendix:proof3}
\begin{proof}
We first prove the monotonicity. For any $Q_1, Q_2 \in \gS \times \gA \rightarrow \reals$ such that $Q_1 \geq Q_2$, for any $(s, a) \in \gS \times \gA$, we prove that 
\begin{equation*}
    \gT^*_{\Omega} Q_1 (s, a) \geq \gT^*_{\Omega} Q_2 (s, a).
\end{equation*}
Recall that
\begin{equation*}
\begin{split}
    &\gT^*_{\Omega} Q(s, a) = r(s, a) +
    \\
    & \gamma \lambda \expect_{s^\prime \sim P(\cdot|s, a)} \left[  \log \left( \expect_{ a^\prime \sim \piref (\cdot)} \left[ \exp \left( \frac{Q (s^\prime, a^\prime)}{\lambda}  \right)  \right] \right)   \right].
\end{split}
\end{equation*}
Since $Q_1 \geq Q_2$ and $\lambda > 0$, we have that
\begin{small}
\begin{equation*}
    \expect_{ a^\prime \sim \piref (\cdot)} \left[ \exp \left( Q_1 (s^\prime, a^\prime) / \lambda  \right)  \right] \geq
    \expect_{ a^\prime \sim \piref (\cdot)} \left[ \exp \left( Q_2 (s^\prime, a^\prime) / \lambda  \right)  \right].
\end{equation*}
\end{small}
From the monotonicity of the logarithm function, we show that 
\begin{equation*}
    \begin{split}
         &\expect_{s^\prime \sim P(\cdot|s, a)} \left[  \log \left( \expect_{ a^\prime \sim \piref (\cdot)} \left[ \exp \left( \frac{Q_1 (s^\prime, a^\prime)}{\lambda}  \right)  \right] \right)   \right] \geq 
        \\
        &\expect_{s^\prime \sim P(\cdot|s, a)} \left[  \log \left( \expect_{ a^\prime \sim \piref (\cdot)} \left[ \exp \left( \frac{Q_2 (s^\prime, a^\prime)}{\lambda}  \right)  \right] \right)   \right].
    \end{split}
\end{equation*}
Then it is direct to obtain that for any $(s, a) \in \gS \times \gA$, $\gT^*_{\Omega} Q_1 (s, a) \geq \gT^*_{\Omega} Q_2 (s, a)$. 

Next we prove that $\gT^*_{\Omega}$ is a $\gamma$-contraction. For any $Q_1, \; Q_2 \in \gS \times \gA \rightarrow \reals$,
$
    \| \gT^*_{\Omega} Q_1 - \gT^*_{\Omega} Q_2 \|_{\infty} = \max_{(s, a) \in \gS \times \gA} \vert \gT^*_{\Omega} Q_1 (s, a) - \gT^*_{\Omega} Q_2 (s, a) \vert.
$
Given $(s, a) \in \gS \times \gA$, without loss of generality, we assume that $\gT^*_{\Omega} Q_1 (s, a) \geq \gT^*_{\Omega} Q_2 (s, a)$. Then we have $\vert \gT^*_{\Omega} Q_1 (s, a) - \gT^*_{\Omega} Q_2 (s, a) \vert = \gT^*_{\Omega} Q_1 (s, a) - \gT^*_{\Omega} Q_2 (s, a)$. Let  $
    \pi_1 (\cdot|s) = \argmax_{\pi} \bigg\{ \expect_{a \sim \pi (\cdot|s)} \left[ Q_1 (s, a) \right] - 
    \lambda \KL \left( \pi (\cdot|s), \piref (\cdot) \right) \bigg\}.
$
and 
$
    \pi_2 (\cdot|s) = \argmax_{\pi} \bigg\{ \expect_{a \sim \pi (\cdot|s)} \left[ Q_2 (s, a) \right] - 
    \lambda \KL \left( \pi (\cdot|s), \piref (\cdot) \right) \bigg\}.
$
Then, we show that
\begin{equation*}
\begin{split}
    &\gT^*_{\Omega} Q_1 (s, a) = r(s, a) + \gamma \expect_{s^\prime \sim P(\cdot|s, a)} \left[ V_1 (s^\prime) \right], \text{with}
    \\
    & V_1 (s) = \expect_{a \sim \pi_1 (\cdot |s)} \left[ Q_1 (s, a) \right] - \lambda \KL \left(\pi_1(\cdot|s), \piref(\cdot)  \right),
\end{split}
\end{equation*}
and 
\begin{equation*}
\begin{split}
    &\gT^*_{\Omega} Q_2 (s, a) = r(s, a) + \gamma \expect_{s^\prime \sim P(\cdot|s, a)} \left[ V_2 (s^\prime) \right], \text{with}
    \\
    & V_2 (s) = \expect_{a \sim \pi_2 (\cdot |s)} \left[ Q_2 (s, a) \right] - \lambda \KL \left(\pi_2 (\cdot|s), \piref(\cdot)  \right).
\end{split}
\end{equation*}
Then, we obtain that
\begin{equation*}
    \begin{split}
        &\quad \gT^*_{\Omega} Q_1 (s, a) - \gT^*_{\Omega} Q_2 (s, a)
        \\
        &=\gamma \expect_{s^\prime \sim P(\cdot|s, a)} \left[ V_1 (s^\prime) - V_2 (s^\prime) \right]
        \\
        &\overset{(1)}{\leq} \gamma \expect_{s^\prime \sim P(\cdot|s, a)} \big[ \expect_{a \sim \pi_2 (\cdot |s^\prime)} \left[ Q_1 (s^\prime, a^\prime)  \right] -
        \\
        &\quad \lambda \KL \left(\pi_2 (\cdot|s^\prime), \piref(\cdot) \right)
         - V_2(s^\prime)    \big]
        \\
        &= \gamma \expect_{s^\prime \sim P(\cdot|s, a), a \sim \pi_2 (\cdot |s^\prime)}  \left[ Q_1(s^\prime, a^\prime) -  Q_2 (s^\prime, a^\prime)  \right]
        \\
        &\leq \gamma \|Q_1 - Q_2 \|_{\infty}.
    \end{split}
\end{equation*}
Inequality $(1)$ holds since 
\begin{small}
\begin{equation*}
    V_1(s^\prime) = \max_{\pi} \bigg\{ \expect_{a \sim \pi (\cdot|s)} \left[ Q_1 (s, a) \right] - 
    \lambda \KL \left( \pi (\cdot|s), \piref (\cdot) \right) \bigg\}.
\end{equation*}
\end{small}
Thus, we prove that the regularized Bellman optimality operator is a $\gamma$-contraction.
\end{proof}

\begin{table*}[htbp]
\small
    \caption{A summary of the comparison results between \methodname~and baseline RL methods in three sparse action tasks. Each data is averaged over the episodic reward of the last five checkpoints.}
    \centering
    \begin{tabular}{c|c|c|c|c|c|c|c|c|c}

    \toprule
    \diagbox{task}{method} & \methodname & PPO & SAC & DQN & ICM & SDQN & RAC-exp & RAC-cos & RAC-tsallis \\   
    \midrule
    Stock & \textbf{4.721}  & 0.256  & 0.720  & 0.0  & -1.173  & -1.332 & -0.104  & -0.186 & -0.540 \\
    \midrule
    Gunplay & \textbf{4.408} & 2.758  & 1.562  & -0.584 & 1.915 & 0.807 & 0.397 & 0.441 & 0.772 \\
    \midrule
    Football & \textbf{0.568}  & 0.418  & -1.0  & -0.995  & -0.965 & -0.946 & -1.000 & -0.996 & -1.000 \\
    \bottomrule
    \end{tabular}
    \label{exptab:all_exp_baselines}

\end{table*}
\vspace{-1em}

\subsubsection{Proof of Proposition \ref{proposition:lower_bound}}
\label{appendix:proof4}

\begin{proof}
First, we show that for any policy $\pi$ and state $s$,
\begin{equation*}
    0 \leq \KL \left( \pi (\cdot|s), \piref (\cdot) \right) \leq \max_a \log(1/\piref(a)).
\end{equation*}

The first inequality holds as KL divergence is non-negative. For the second inequality, we have that
\begin{equation*}
    \begin{split}
        &\quad\KL \left( \pi (\cdot|s), \piref (\cdot) \right)
        \\
        &= \sum_{a \in \gA} \pi (a|s) \log \left( \pi (a|s) \right) +  \sum_{a \in \gA} \pi (a|s) \log \left( 1 / \piref (a) \right)
        \\
        &\overset{(1)}{\leq} \sum_{a \in \gA} \pi (a|s) \log \left( 1 / \piref (a) \right) 
        \\
        &\overset{(2)}{\leq} \max_a \log(1/\piref(a)).
    \end{split}
\end{equation*}

Inequality $(1)$ holds from the non-negativity of entropy. Inequality follows as $0 \leq \pi (a|s) \leq 1, \; \forall a \in \gA$. Let $C = \lambda \max_a \log(1/\piref(a))$. Then we show that for any policy $\pi$,
\begin{equation}
\label{equation:proof}
   V^{\pi} (s) - \frac{C}{1-\gamma} \leq V^{\pi}_{\Omega} (s) \leq V^{\pi} (s), \quad \forall s \in \gS.
\end{equation}
By definition, we have 
\begin{small}
\begin{equation*}
    V^{\pi}_{\Omega} (s) = \expect \bigg[ \sum_{t=0}^\infty \gamma^t \left( r (s_t, a_t) - \lambda \KL \left( \pi (\cdot|s_t) , \piref (\cdot) \right) \right) \big| s_0 = s \bigg].
\end{equation*}
\end{small}

It is direct to show that $V^{\pi}_{\Omega} (s) \leq V^{\pi} (s)$. Furthermore, we show that
\begin{small}
\begin{equation*}
    \begin{split}
        V^{\pi}_{\Omega} (s) &= V^{\pi} (s) - \lambda \expect \bigg[ \sum_{t=0}^\infty \gamma^t  \KL \left( \pi (\cdot|s_t) , \piref (\cdot) \right) \big| s_0 = s \bigg]
        \\
        &\geq V^{\pi} (s) - \frac{C}{1-\gamma}.
    \end{split}
\end{equation*}
\end{small}

Then, we show the sub-optimality of $\pi^*_{\Omega}$.
\begin{small}
\begin{equation*}
    V^{\pi^*_\Omega} (s) \overset{(1)}{\geq} V^{\pi^*_\Omega}_{\Omega} (s) \overset{(2)}{\geq} V^{\pi^*}_{\Omega} (s) \overset{(3)}{\geq} V^{\pi^*} - \frac{C}{1-\gamma} = V^{*} - \frac{C}{1-\gamma}.
 \end{equation*}
\end{small}
Inequality $(1)$ and $(3)$ follow Eq. (\ref{equation:proof}). Inequality $(2)$ holds as $\pi^*_{\Omega}$ is the regularized optimal policy. Therefore, we finish the proof.
\end{proof}

\subsection{Additional Results}
\label{subsec:additional_experiments}

\begin{figure}[htbp]
  \centering
  \subfigure{
      \includegraphics[width=1\linewidth]{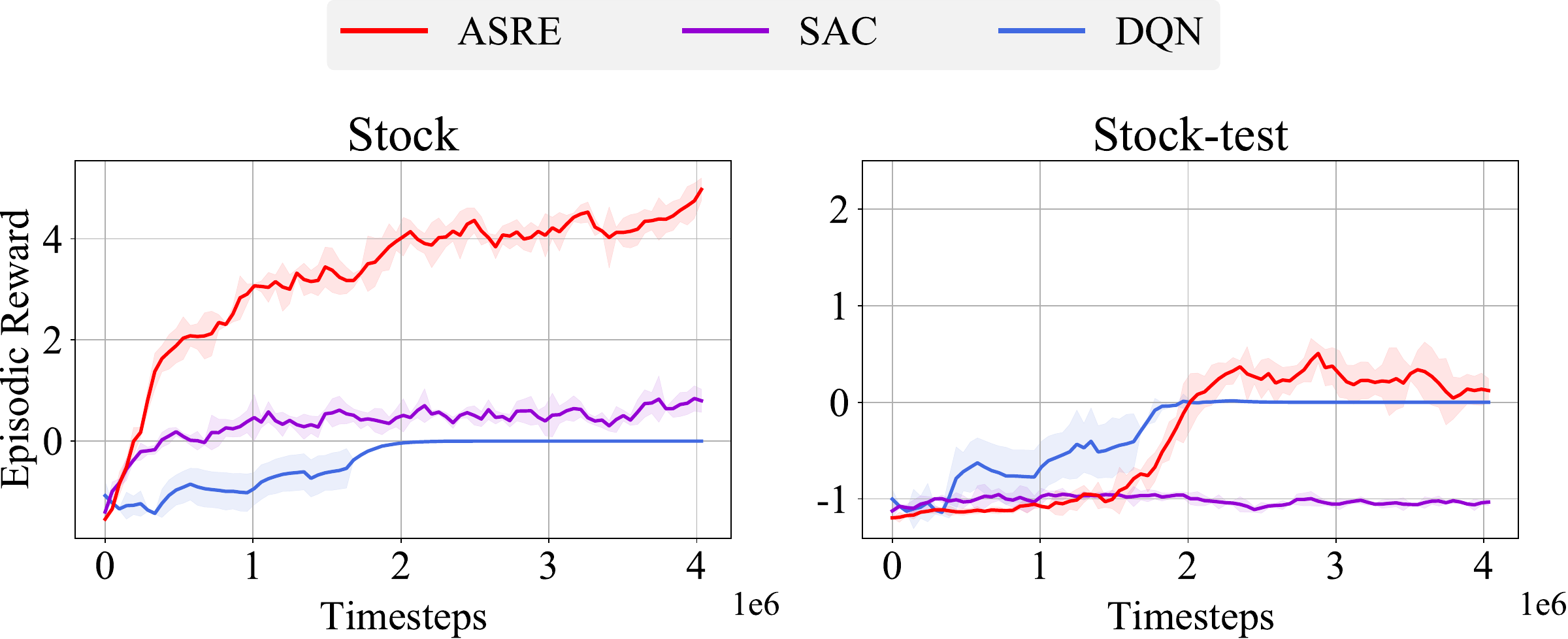} 
  }
  \caption{Training curves on training and testing stock dataset.}
  \label{expfig:split_stock_dataset}
\end{figure}

\textbf{Evaluation on testing stock data}. 
In the main experiments, we present the results only on the training stock data. Here we consider a more realistic setting: we partition the stock dataset into a training set (data spanning January 10, 2017, to April 28, 2017) and a testing set (data covering April 29, 2017, to May 26, 2017). The policy is only trained on the training set and evaluated on both the training and testing sets. Fig. \ref{expfig:split_stock_dataset} illustrates the training process for the training and testing sets (denoted as Stock and Stock-test, respectively). \methodname~achieves a positive score on the Stock-test, indicating its potential for practical application. DQN converges at a reward of 0, suggesting that the policy trained by DQN may be overly conservative and abstain from making any trades.

\begin{figure}[htbp]
  \centering

  \subfigure{
      \includegraphics[width=1\linewidth]{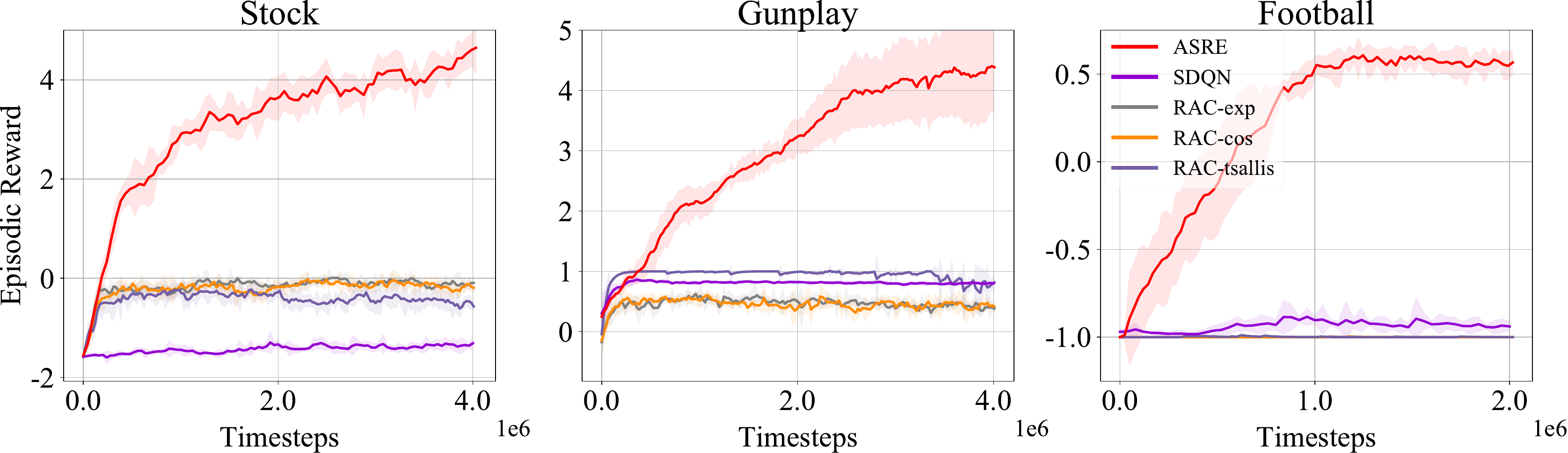} 
  }

  \caption{Training curves of \methodname~and regularization-based RL methods in sparse action tasks.} 
  \label{expfig:full_results_regularizer}
\end{figure}

\textbf{Comparison with regularized-based method}. \methodname~is a regularized-based RL method that proposes to explore according to action sparsity. We conduct experiments comparing \methodname~to regularization-based baselines and exploration RL method: Sparse Deep Q Networks (SDQN~\citep{SDQN}) and Regularized Actor Critic (RAC~\citep{RAC_regularized_actor_critic}) are RL methods that regularize reward functions with various regularizers. Their regularizers include Tsallis entropy ($\frac{1}{2}(1-x)$), exp function ($\exp(1)-\exp(x)$) and cos function ($\cos(\frac{\pi}{2}x)$). Fig. \ref{expfig:full_results_regularizer} shows the comparisons between \methodname~and regularization-based methods. These common regularizers do not consider the sparsity of the actions. Therefore, they do not help improve policy learning.

\begin{figure}[htbp]
  \centering
  \subfigure{
      \includegraphics[width=1\linewidth]{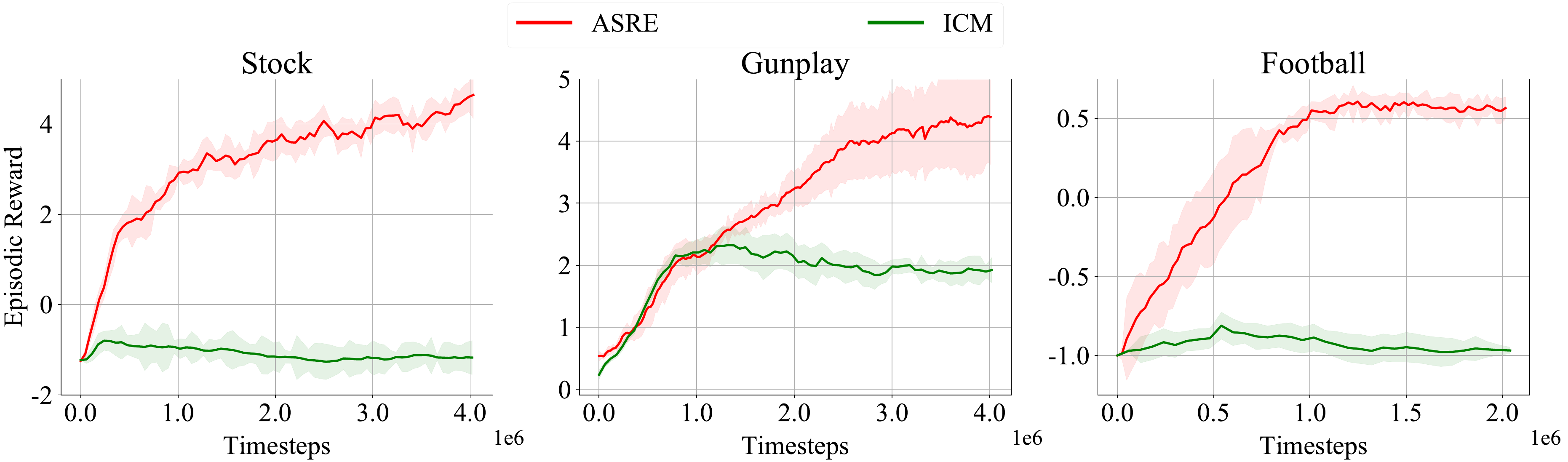} 
  }
  \caption{Training curves of \methodname~and ICM.} 
  \label{expfig:full_results_exploration}
\end{figure}

\textbf{Comparison with exploration method in RL}. We also compare \methodname~to Intrinsic Curiosity Module (ICM~\citep{ICM}), a classic exploration technique in RL. Fig. \ref{expfig:full_results_exploration} shows the comparison result. Regarding policy improvement speed and final score, ICM cannot learn in sparse action tasks efficiently. This is because the curiosity-based method focuses on exploring unseen states, ignoring the sparsity property of sparse action, and wastes the budget for executing sparse action. This result demonstrates that the curiosity-based exploration technique widely employed in the current RL community is ineffective for sparse action tasks.

We summarize the comparison results between \methodname~and all baselines in Tab. \ref{exptab:all_exp_baselines}. In sparse action tasks, \methodname~can outperform all these RL baselines that have been widely applied to various RL tasks. 

\begin{figure}[t]
    \centering
    \includegraphics[width=1\linewidth]{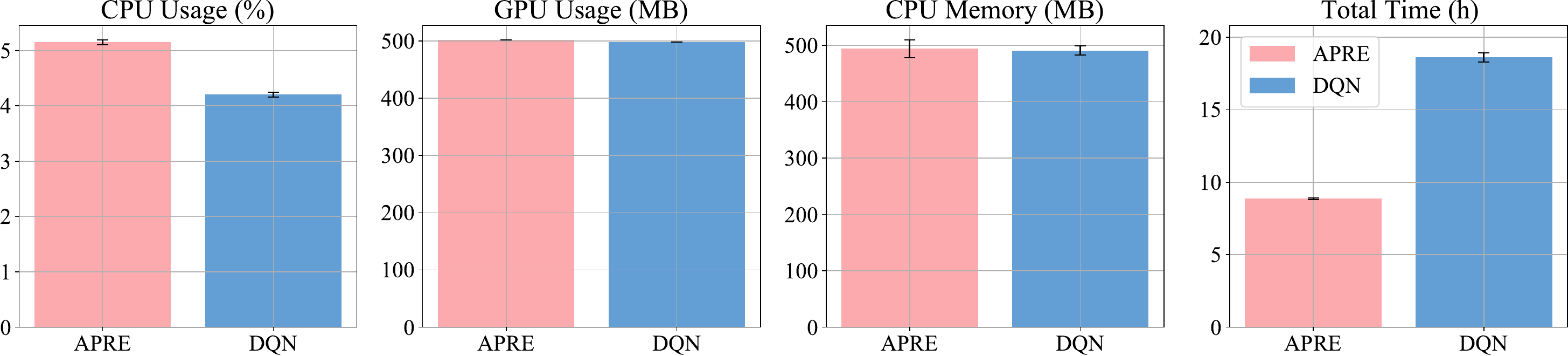}
    \caption{Computational efficiency of \methodname~and DQN method. \methodname~consumes comparable GPU and CPU resources, but significantly saves time budget and improves performance, compared to DQN. Error bars represent standard deviation across five runs.}
    \label{fig:efficacy}
\end{figure}

\textbf{Computational efficiency.} We conduct an analysis to validate the impact of \methodname~method on computational resources and time efficiency. We record various computational metrics during the running of \methodname~and classic RL algorithm, DQN. The metrics include CPU and GPU utilization, memory consumption, and the total elapsed time. These metrics are represented in Fig. \ref{fig:efficacy}. The results indicate that \methodname~maintains a comparable level of CPU and GPU usage to that of the DQN algorithm. However, it is paramount to highlight that \methodname~significantly reduces the time required to achieve the learning objectives, thereby adhering to the time budget constraints and enhancing learning performance.

\subsection{Implementation Details}

Our algorithm \methodname~is implemented based on \textit{stable-baselines} \citep{stable-baselines}. All baseline algorithms are implemented in stable-baselines and other well-verified open-source code repositories. In our experiments, every reported result is evaluated in the environment for 40 episodes and averaged over five runs. Tab. \ref{appendixtable:hyperparameter} presents the hyper-parameters of \methodname~and other baselines used in our experiments. All sparse action tasks are built following SA-MDP. The action execution number limitation $K$ is $30, 5,$ and $4$ for Stock, Gunplay, and Football, respectively, while Atari games are built under common MDP.

\begin{table}[htbp]
\small
    \caption{Hyper-parameters in the experiments.}
    \centering
    \begin{tabular}{c|c}

    \toprule
    \textbf{Hyper-parameter} & \textbf{Value}  \\   
    
    \midrule
    
    \textbf{\methodname} \\
    $\lambda$ & 0.01 \\
    $c$ in D-UCB & 0.5 \\
    $\tilde{\gamma}$ in D-UCB & 0.99 \\
    batch size & 256 \\
    learning rate & 0.001  \\ 
    buffer size & 400000\\
    Q network & Adam, [256,128], relu \\
    N for sparsity evaluation & 30 \\
    target smoothing ratio & 0.005 \\
    $\delta$ for constraining sampling & $1/(8|\gA|)$ \\ 
    
    \midrule
    \textbf{DQN} \\
    batch size & 256 \\
    learning rate & 0.001 \\    
    buffer size & 400000\\
    Q network & Adam, [256,128], relu \\
    practical tricks [1] & double, noise layer \\
    practical tricks [2] & dueling, prioritized \\
    
    \midrule
    \textbf{SAC} \\
    batch size & 256 \\
    learning rate & 0.001 \\
    target smoothing ratio & 0.005 \\
    buffer size & 400000\\
    regularization coef & 0.05 \\
    Q-network & Adam, [256,128], relu \\

    \midrule
    \textbf{PPO} \\
    clip ratios & 0.2 \\
    epochs & 10 \\
    batch size & 1200 \\
    sample number per update & 12000 \\
    learning rate & 0.0003 \\
    policy network & Adam, [256,128], relu \\
    value network & Adam, [256,128], relu \\
    \bottomrule
    \end{tabular}
    \label{appendixtable:hyperparameter}
\end{table}

\ifCLASSOPTIONcaptionsoff
  \newpage
\fi

\clearpage
\bibliographystyle{IEEEtran}
\bibliography{references}

\end{document}